\documentclass{article}
\usepackage{arxiv}
\usepackage{graphicx}
\usepackage[numbers]{natbib}
\usepackage{doi}
\usepackage[utf8]{inputenc} 
\usepackage[T1]{fontenc}    
\usepackage{url}            
\usepackage{booktabs}       
\usepackage{amsfonts}       
\usepackage{nicefrac}       
\usepackage{microtype}      
\usepackage[table]{xcolor}         
\usepackage{amsmath}
\usepackage{amssymb}
\usepackage{bbm}
\usepackage{pifont}
\usepackage{threeparttable}
\usepackage{subcaption}
\usepackage{array} 
\usepackage{makecell}
\usepackage{algpseudocode} 
\usepackage{multirow}
\usepackage{listings}
\usepackage{enumitem}
\usepackage{hyperref}
\usepackage{todonotes}
\usepackage{comment}
\usepackage{xcolor}

\usepackage{lscape}
\usepackage{tabularray}
\usepackage{CJK}

\usepackage{wrapfig}
\usepackage{float}
\usepackage[ruled,vlined]{algorithm2e}

\newcommand{\eg}{\textit{e.g.}}

\newcommand{\ie}{\emph{i.e.}}

\usepackage{authblk}

\newcommand{\pangu}{Pangu}
\newcommand{\modelname}{Pangu Ultra MoE}
\newcommand{\minisection}[1]{\textbf{#1} }

\title{\modelname{}: How to Train Your Big MoE on Ascend NPUs}

\author{\pangu~Team, Huawei\\
pangutech@huawei.com}

\renewcommand{\headeright}{Technical Report}

\begin{document}

\maketitle

\thispagestyle{fancy}

\begin{abstract}

Sparse large language models (LLMs) with Mixture of Experts (MoE) and close to a trillion parameters are dominating the realm of most capable language models. However, the massive model scale poses significant challenges for the underlying software and hardware systems. In this paper, we aim to uncover a recipe to harness such scale on Ascend NPUs. The key goals are better usage of the computing resources under the dynamic sparse model structures and materializing the expected performance gain on the actual hardware. To select model configurations suitable for Ascend NPUs without repeatedly running the expensive experiments, we leverage simulation to compare the trade-off of various model hyperparameters. This study led to Pangu Ultra MoE, a sparse LLM with 718 billion parameters, and we conducted experiments on the model to verify the simulation results. On the system side, we dig into Expert Parallelism to optimize the communication between NPU devices to reduce the synchronization overhead. We also optimize the memory efficiency within the devices to further reduce the parameter and activation management overhead. In the end, we achieve an MFU of 30.0\% when training Pangu Ultra MoE, with performance comparable to that of DeepSeek R1, on 6K Ascend NPUs, and demonstrate that the Ascend system is capable of harnessing all the training stages of the state-of-the-art language models. Extensive experiments indicate that our recipe can lead to efficient training of large-scale sparse language models with MoE. We also study the behaviors of such models for future reference.
\end{abstract}

\section{Introduction}
Recent advances in sparse large-scale models have positioned Mixture of Experts (MoE) as a key ingredient for Large Language Models (LLMs), thanks to their capacity of learning effectively from tens of trillions of tokens~\cite{sun2024hunyuan, Yang2024Qwen25TR, jiang2024mixtral, deepseekv3, li2025minimax, Llama42025, deepseekai2025deepseekr1incentivizingreasoningcapability}. The inference of MoE models is also more efficient than that of dense models because only a fraction of parameters are activated for a token~\cite{shazeer2017outrageously}.
However, training such large MoE models is no small feat, which requires sustained orchestra of thousands of AI computing nodes~\cite{deepseekai2025deepseekr1incentivizingreasoningcapability}. In addition, although the sparse structure can theoretically reduce the required computation by an order of magnitude, materializing the reduction is challenging because the actual fraction of parameters is dynamically determined jointly by the input tokens and the parameter states. Therefore, the computing cluster systems usually suffer from inefficient utilization when training large-scale MoE models.

In this report, we aim to improve system utilization to make it effective and efficient to train large MoE language models on Ascend NPUs. Our design consists of two key aspects: model architecture design and system optimization. We hope to find the right model configuration that can be efficiently supported on Ascend NPUs, while reducing the overhead of communication between memory and processors as well as between NPUs.

We first examine systematic architectural exploration to look for efficient model configurations on Ascend NPUs. We employ a two-level approach: pilot experiments for the MoE block design and system simulation for optimal model structures on Ascend NPUs. First, we conduct pilot studies on two critical aspects of MoE models: expert granularity~\cite{dai2024deepseekmoe} and the number of shared experts~\cite{lepikhin2020gshard, liu2024deepseek}. Then, we develop a simulator to predict model performance by analyzing model throughput. The prediction employs a bottom-up workflow, with validating individual operators on Ascend NPUs as simulation basis, then to evaluate end-to-end system efficiency. To improve the precision of our simulation, we also consider computation-communication patterns across layers, operator-level interactions, and hardware-specific constraints. This simulation gives us the configuration of Pangu Ultra MoE, a sparse LLM with 718 billion parameters. During the actual model training, we also examine expert load imbalance and token drop strategy under the setup of the large MoE model with close to one trillion parameters. We find its training behavior is different from smaller sparse models.

We then try to improve system utilization when training the models on thousands of Ascend NPUs. When conducting such experiments on large-scale MoE models, we encountered three bottlenecks: device
communication overhead, high memory utilization, and imbalanced expert load. Traditional All-to-All communication in expert parallelism does not distinguish intra-node and inter-node traffic~\cite{megatron-lm}, leading to suboptimal bandwidth utilization and increased communication overhead. Most widely used memory-saving strategy recomputation focuses on the entire layer or the whole module~\cite{megatron-sp}, failing to effectively balance memory and runtime efficiency. Moreover, although auxiliary loss~\cite{fedus2022switch} may implicitly control the degree of expert balance through regularization, they do not adapt well to real-time fluctuations in computational load across devices.

To address those challenges accordingly, we first carry out a fine-grained parallelism strategy and optimize communication patterns. To improve the efficiency of All-to-All communication in Expert Parallelism (EP), we introduce Hierarchical EP All-to-All Communication, optimizing bandwidth utilization by separating inter-node Allgather and intra-node All-to-All communications. These two communications are both effectively overlapped with computations by our proposed Adaptive Pipe Overlap mechanism, which leverages finer-grained operations to overlap communication with computation. To ensure stability within limited NPU memory, we then improve the efficiency of memory utilization by two techniques. First, we conduct fine-trained recomputation that selectively recomputes intermediate activations of specific operators instead of storing all activations or recomputing the entire layer. Second, we propose tensor swapping that offloads activations to the host temporarily and prefetches them for backward computation, without storing them on the NPU. To mitigate expert load imbalance, we propose a dynamic device-level load balancing approach using real-time load expert prediction and adaptive expert placement across NPUs.

Taking advantage of our model selection strategy and parallel computing system optimization, when training \modelname, with performance comparable to DeepSeek R1~\cite{deepseekai2025deepseekr1incentivizingreasoningcapability}, we achieve a Model Flops Utilization (MFU) of 30.0\% and Tokens Per Second (TPS) of 1.46M on 6K Ascend NPUs, compared to the baseline MFU of 18.9\% and TPS of 0.61M on 4K Ascend NPUs. In particular, \modelname~performs well on medical benchmarks. We also study the model behaviors, especially the MoE structure, to uncover further guidelines for training large-scale MoE models on Ascend NPUs.

\section{MoE Architecture Design for Ascend NPUs}

Designing an optimal model architecture for a new hardware platform typically involves extensive trial and error, which becomes prohibitively expensive for large-scale MoE models. Given the vast search space of software and hardware configurations, including model parameters, NPU capabilities, training efficiency, and inference performance, a rapid exploration method with quantitative estimations is essential~\cite{10.1145/3581784.3607102}. To address this, we propose an efficient, simulation-based approach for architecture search. Our method first determines the MoE structure through reduced-scale training experiments before finalizing the full model. By combining this with analytical pre-pruning to narrow the search space, we derive a set of candidate architectures optimized for Ascend NPUs, validated via a high-level simulation framework. Finally, we evaluate the suitability of the resulting architecture for Ascend NPUs.

\subsection{MoE Block Design}
\label{sec:moe_design}

To guide our model design, we first analyze contemporary MoE architectures~\cite{deepseekv3,muennighoff2024olmoe,li2025minimax}, systematically evaluating two key dimensions: expert partitioning granularity and the role of shared experts. To establish empirical insights and derive scalable design principles, we develop and assess a smaller-scale MoE model as a preliminary baseline. This model consists of 256 total experts, activates \texttt{topk=8} experts per token (including a default shared expert unless specified), and contains approximately 20B total parameters, with 1.6B activated per token.

\begin{figure}[t]
    \centering
    \begin{minipage}[b]{0.47\textwidth}
        \centering
        \includegraphics[width=\textwidth]{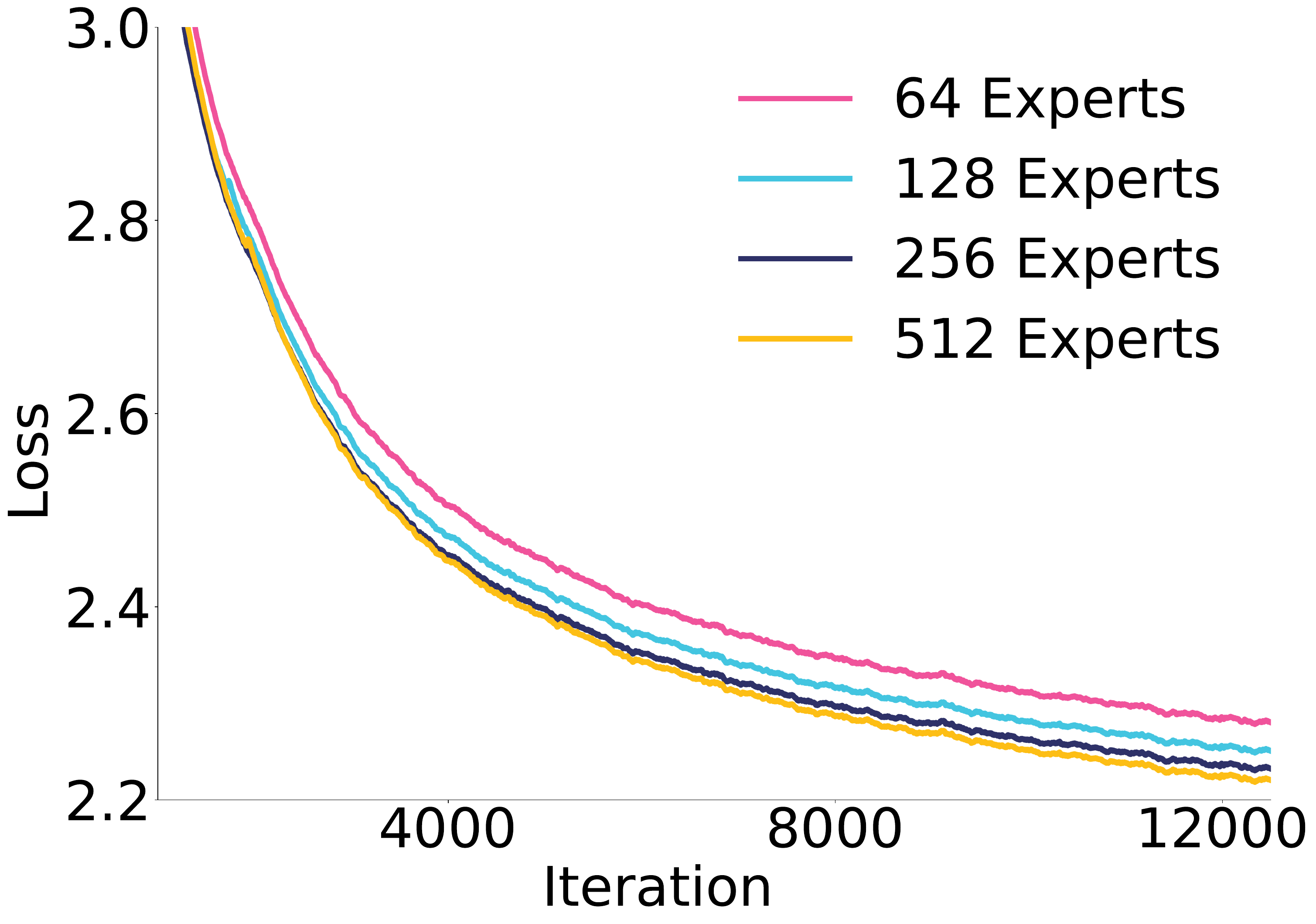}
        \caption{Experts number ablation. Since increasing experts from 256 to 512 brings limited gains, we select 256 experts for \modelname~ to balance performance and practical efficiency.}
        \label{fig:num_expert_ablation}
    \end{minipage}
    \hfill
    \begin{minipage}[b]{0.4\textwidth}
        \centering
        \includegraphics[width=\textwidth]{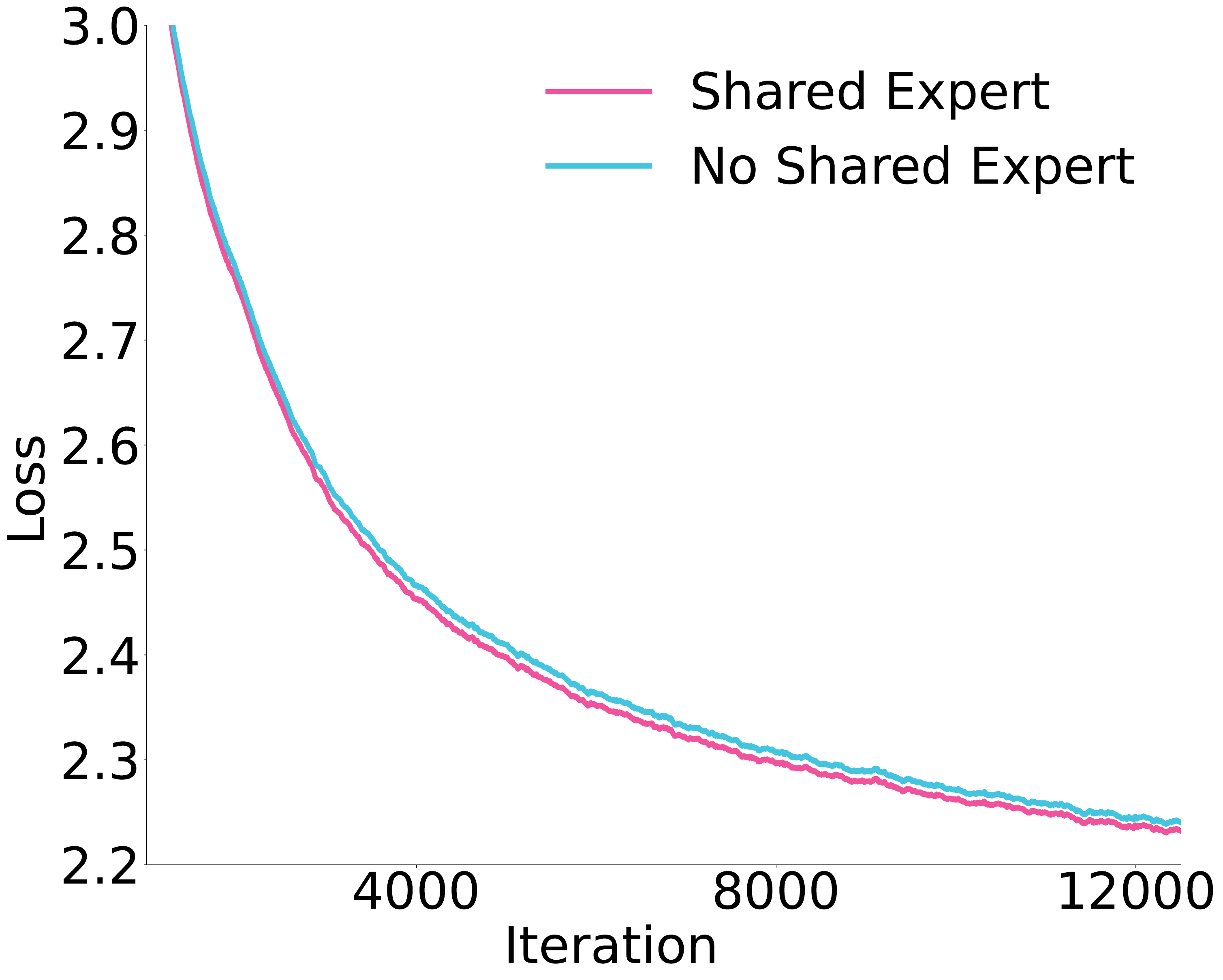}
        \caption{Shared expert ablation. We compare the 20B baseline MoE (\texttt{topk=8}) and its no shared expert variant (\texttt{topk=9}). The former leads to slightly lower training loss.}
        \label{fig:shared_expert_ablation}
    \end{minipage}
    \label{fig:moe_ablations_combined}
\end{figure}

\minisection{Expert Granularity}
The granularity of expert partitioning, balancing the number of experts against their individual size, critically influences model performance, efficiency, and resource utilization. Recent work \cite{dai2024deepseekmoe} has systematically demonstrated that expanding the expert population while maintaining a fixed per-token computational budget can substantially improve model expressiveness. Our experiments corroborate these findings: when keeping the activated parameters per token constant, configurations with 256 experts achieve significantly lower training loss compared to those with 64 or 128 experts (Figure~\ref{fig:num_expert_ablation}). However, further scaling to 512 experts yields diminishing returns, with marginal improvements over the baseline of 256 experts. This suggests a saturation effect where simply increasing expert count provides limited benefits under fixed computational constraints. Balancing these empirical observations against practical considerations of training/inference performance and resource efficiency, we identify 256 experts as a good trade-off point. This observation aligns with the architectural analysis of Huang et al.~\cite{huang2024harder}, which demonstrates that increasing expert diversity through larger expert pools enhances task specialization without increasing computational costs.

\minisection{Shared Expert Architecture}
Recent works have explored the design and effectiveness of shared experts within MoE architectures. In particular, incorporating shared experts has been shown to reduce training costs and improve inference efficiency~\cite{qwen_moe}. The foundational works on MoE in Transformers, such as GShard~\cite{lepikhin2020gshard}, introduced concepts like expert sharding and conditional computation, which can be seen as precursors to modern shared expert designs. To further evaluate the efficiency of the shared expert, we compare our baseline configuration against a variant with the same total number of experts but activating \texttt{topk=9} experts and adopting no shared expert. This design ensures a comparable number of activated parameters per token, as both configurations activate 9 experts during each forward pass. As illustrated in Figure~\ref{fig:shared_expert_ablation}, the model employing shared experts achieves a lower training loss than its counterpart without shared experts. Based on these results,  we adopt the shared expert architecture for our final model.

\subsection{Architecture Design Optimized for Ascend NPUs}
\label{section: Simulation}

\minisection{Architecture Design Space} Considering the computational workload, we expect the MoE model will have a total parameter count of approximately 700 billion. In terms of the depth-to-width ratio, as explored in the previous studies~\cite{kaplan2020scaling, levine2020depth}, the optimal depth-to-width recommendation from empirical LLM practice is fitted as:
\begin{equation*}
    \log(d_{\mathtt{hidden}})=(5.039\pm 0.030)+(5.55 \times 10^{-2}\pm 1.3 \times 10^{-3})L,
    \label{eq:depth-to-width ratio}
\end{equation*}
where $d_{\mathtt{hidden}}$ represents the hidden dimension, and $L$ denotes the number of transformer blocks. Based on this equation and considering the MoE granularity, we perform an extensive exploration of the parameter space, which includes configurations such as Multi-Head Latent Attention (MLA) \cite{liu2024deepseek} configurations, Feedforward Network (FFN) dimensions, expert dimensions, the number of experts, and activation ratios, among other factors. This comprehensive analysis allows us to define feasible parameter ranges, effectively narrowing the search space to approximately 10,000 unique configurations.

\begin{figure*}[t]
    \centering
    \includegraphics[width=0.75\textwidth]{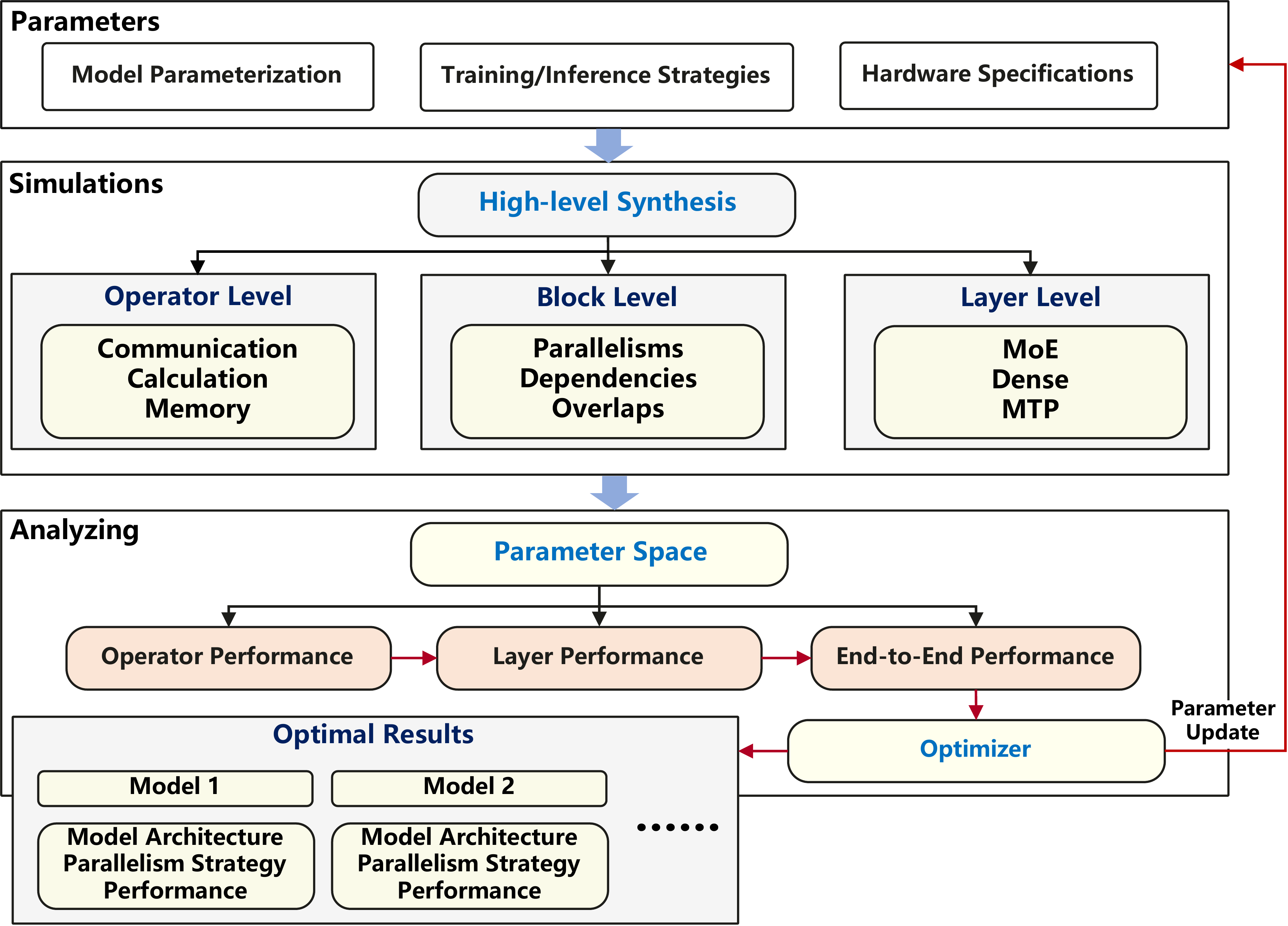}
    \caption{Simulation workflow for LLM performance and optimal search.}
    \label{fig:simulation_workflow}
\end{figure*}

\minisection{Architecture Search via Simulation} We aim to conduct simulations to explore the optimal architecture within any given parameter search space. In general, the parameter space for our high-level simulation method encompasses the model architecture, the number of weights,  possible parallelism strategies, and hardware system descriptions. Representative values such as peak floating-point operations per second (FLOPS), communication and memory bandwidth, on-chip memory, and cluster specifications are used to characterize the computing power. Our MoE model consists of dense, MoE, and MTP layers, each differing in computational and communication load on the hardware system. Hyperparameters, including the number of layers, hidden sizes in both the dense and MoE components, and the number of experts, are all considered in the simulation. We also explore a 5D parallelism strategy: Pipeline Parallelism (PP)~\cite{huang2019gpipeefficienttraininggiant} (including Virtual Pipeline Parallelism, VPP~\cite{narayanan2021efficientlargescalelanguagemodel}), Tensor Parallelism (TP)~\cite{shoeybi2020megatronlm}, Expert Parallelism (EP)~\cite{lepikhin2020gshard}, Data Parallelism (DP)~\cite{rajbhandari2020zero} and Context Parallelism (CP)~\cite{jacobs2023deepspeedulysses}. These strategies are combined to identify the resource allocation that achieves the best performance. The detailed computation and communication process consists of a series of computational or communicational operations (\eg, GEMM, Reduce-Scatter) across the layers. This includes memory access and utilization from various memory tiers, as well as computation-communication overlap, recomputes,  and memory offloading when necessary. The time required for each given data batch is predicted, which can be directly converted into throughput and MFU. To provide a broad overview of the simulation workflow instead of exhaustively listing all the relevant parameters, the mainstream simulation process, as shown in Figure~\ref{fig:simulation_workflow}, is applied to both the training and inference processes, with the caveat that training involves backward propagation, which introduces additional computational and memory overhead.

For each architecture, we evaluate the impact of different parallelism strategies. To accurately simulate model performance on the Ascend 910B platform, we model all computations, communications, and memory accesses in the network, taking into account parameters such as computational power, bandwidth, latency, and efficiency. Additionally, each network is characterized by its handling of specific operations, including overlapping with computational operators, steady stage in pipeline, TP, EP, etc. By intensively considering all possible features and model architectures, we maximize the utilization of the Ascend 910B platform, mitigating computational bottlenecks through strategic combinations of strategies.

\subsection{Results of Architecture Search} Prior to conducting architecture search via simulation, we first validate the simulation accuracy by conducting experiments on MoE models for end-to-end duration testing in training. Once the simulation method has been validated, an optimal model architecture is proposed for our \modelname{}. 

\minisection{Validations of the Simulation Method}
Our simulation method integrates comprehensive hardware characteristics in the physical process, including Cube/Vector compute capabilities, memory access performance, and network communication bandwidth, to enable fine-grained modeling of system behavior. Since hardware performance heavily depends on input-dependent factors (\eg, tensor shapes, communication domains, and data transfer volumes), precision evaluations requires aligning the configuration parameters of simulation with the real-world hardware executions. 

To achieve this, we conduct 2 training experiments on our Ascend NPU clusters and perform corresponding simulation with exactly the same configurations. Firstly, we build a 6-layer MoE model with totally 4.2B parameters for training experiment on 128 Ascend NPUs. This model adopts grouped-query attention~\cite{ainslie2023GQA} with 96 heads, 96 KV heads, and 12288 hidden size. Each layer of the model has 4 experts with expert dimension of 49152 while 2 of the experts are activated for each token. During the training, parallelism strategy is set as TP = 8, EP = 4, DP = 16 and PP = 1. The sequence length is 8192, global-batch-size is 64 and micro-batch-size is 1. The end-to-end time in the second experiment with 718B experiments is acquired after actual \modelname{} training on 6K NPUs, and we carry out simulations afterward to re-validate our simulation method. We include this result here to further demonstrate the effectiveness of our simulation. 

The simulated and measured results of end-to-end durations are summarized in Table \ref{tab:Comparison of Operator-level and End-to-End Duration}. The simulation achieves satisfactory accuracy, exceeding 85\% in both experiments, proves the scalability and accuracy of end-to-end durations for computation-intensive workloads. However, the simulated computation and end-to-end durations are consistently shorter than the measured data. This discrepancy arises from idealized assumptions, such as perfectly distributed computations, ideally overlapped communications, and efficient memory access across NPUs, while neglecting unpredictable factors like cache hit rates. Despite these limitations, the end-to-end accuracy reaches over 85\%, indicating that the simulation accurately models the physical process and reliably produces valid results.

\begin{table}[t] 
\centering 
\caption{Comparison of measured and simulated time for one complete forward and backward pass.} 
\label{tab:Comparison of Operator-level and End-to-End Duration} 
\begin{tabular}{cccc} 
    \toprule 
    & Measured Duration (s) & Simulation Results (s) & Accuracy  \\ 
    \midrule
     \makecell[c]{Experiment on 128 NPUs\\(4.2B Model Preliminary Experiment)} & 3.40 & 3.03 & 88.9\%   \\
      \midrule
     \makecell[c]{Experiment on 6K NPUs\\(\modelname{} Training)}
      & 17.20       & 15.49        & 90.1\% \\
     
    \bottomrule 
\end{tabular} 
\end{table}

\minisection{Analysis of Selected Models} We evaluate the model performance under all parameter combinations via simulation using the workflow as in Figure \ref{fig:simulation_workflow}. By analyzing both training and inference throughput of all the models in the search space of around 10,000 configurations, we identify model architectures with the highest performance. Table \ref{tab:Top-8 Model Architectures}~lists the comparison of parameters between the 8 most advantageous samples. Apart from the differences listed, all 8 models adopt typical MLA \cite{liu2024deepseek} with 128 heads. Each MoE block has 1 shared expert and varying number of total routed experts (but always 8 activated). The intermediate size of both shared expert and routed experts is 2048.

\begin{table}[htbp] 
\centering 
\caption{Comparison of 8 model architectures with highest performance identified by simulation evaluations.} 
\label{tab:Top-8 Model Architectures} 
\begin{tabular}{lcccccccc} 
    \toprule 
    Model & 1 & 2 & 3 & 4 & 5 & 6 & 7 & 8 \\ 
    \midrule 
    Number of Layers & 61 & 62 & 66 & 70 & 70 & 70 & 61 & 66 \\ 
    Model Hidden Size & 7168 & 7168 & 7168 & 6144 & 7168 & 7168 & 7680 & 7680\\
    Number of Routed Experts & 256  & 256 & 240 & 256 & 240 & 256 & 256 & 256 \\
    \bottomrule 
\end{tabular} 
\end{table}

\begin{figure*}[ht!]
    \centering
    \includegraphics[width=0.6\textwidth]{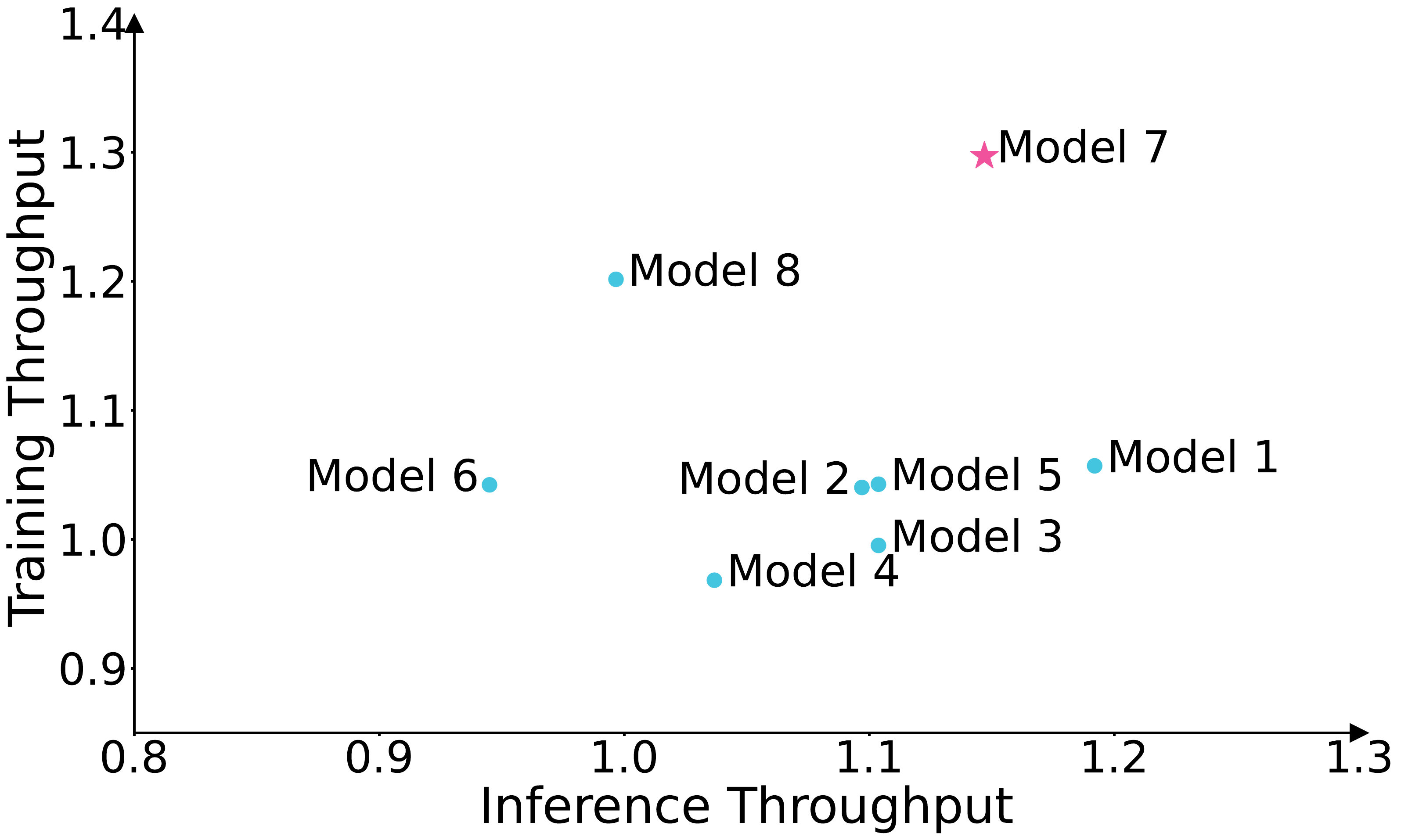}
    \caption{Simulated model normalized throughput of training and inference.}
    \label{fig:throughput}
\end{figure*}

The relative comparisons of simulated model training and inference throughput of the highest 8 models are illustrated in Figure~\ref{fig:throughput} for detailed analysis. We have proved that our simulated data tend to show slightly shorter end-to-end time than reality, which turns out to be moderately overestimated throughput in all models. Despite this, as we are analyzing comparative optimal throughput, the overall higher values in a similar extent will not affect the comparative result.

Regarding our architectural design, the hidden size and the number of layers show significant impacts on model performance. Model 7 with 61 layers reaches nearly 15\% higher training throughput than model 8 with 66 layers, while all other parameters remain unchanged. Reducing the model hidden size (like model 7 vs. model 1, and model 5 vs. model 4) will also cause degraded performance by around 4\% to 25\%. Besides, increasing the number of routing experts introduces additional computational overhead due to dynamic routing decisions and communication costs, leading to severe inference throughput degradation (Model 6 is 14.4\% lower than Model 5).

Models in Figure~\ref{fig:throughput} show significantly different performance in training and inference. In most cases, training tends to be computation-bounded while inference is memory access bounded. The relative difference between training and inference is generally related to the workload of computations and volume of activations of each model. Model 7 with 61 layers and 7680 hidden size, as labeled by the pentagram in the Figure, shows superior training and inference throughput.

In conclusion, by thoroughly analyzing all the candidates with simulations, model 7 significantly outperforms all the other models in terms of overall performance. Our simulation also suggests a combination of optimized parallelism strategies for this model to achieve the highest throughput: TP = 8, PP = 16, VPP = 2, EP = 4, and MBS = 2. We adopt this parallelism strategy for our actual \modelname{} model training, while its implementations will be introduced in Section~\ref{section: Parallelization Strategy Optimization}. We attribute the high throughput of model 7 to its highly matched architectures with our Ascend NPU clusters. The detailed analysis of its optimal number of experts, hidden size, and number of layers is further demonstrated in the next section.

\subsection{Conclusion on Architecture Design Optimized for Ascend} 

 In the process of architecture design, we first ensure the effectiveness of MoE structure for stable model outcomes as a starting point, then we create an overall design space of all available combinations of the rest of model hyperparameters under computational constrains for simulation search toward optimized throughput. By analyzing the throughput outcomes of all the models in the search space, we notice the selection of hyperparameters like the number of experts, model hidden size and number of layers should be considered along with Ascend hardware features and model parallelisms. We concludes the crucial factors as below:

\minisection{Computation-Communication-Memory Balance} An optimal model on Ascend should be able to balance computing power, communication bandwidth and memory bandwidth based on Ascend system. This prevents any bottleneck from happening during training and inference.

As for training, the hidden size of the model will largely change its computation, communication and memory access pattern, while computation is affected most and memory access least. Our Ascend NPU features a large ratio of computing power to communication or memomy access bandwidth, so models with larger hidden size like model 7 or 8 in Table \ref{tab:Top-8 Model Architectures} are able to improve the utilization of hardware resources, which aligns with our simulated results of training throughput. Differently, as inference is highly memory access required, reducing the volume of activations (like model 1 and 7 with smaller layer numbers) could effectively improve model inference throughput.

Besides, it is worthwhile to mention that although MLA increases computational load, its reduction in KV Cache relieves the pressure on memory usage and bandwidth, offering a higher computation-to-memory access ratio. As a result, MLA is closer to the roofline of Ascend chips than commonly used GQA.

\minisection{Expert Number vs.\ Parallelism}
MoE structure greatly expands the number of parameters but also requires large amount of memory usage and communication. To tackle this challenge on Ascend, we adopt experts with small size and higher sparsity ratio to reduce their workload on Ascend. EP is also an effective tool for reducing memory access and memory usage per NPU, however, an excessively large EP value introduces large communication overhead, causing performance degradations. Besides, applying large TP on granular experts (2048 intermediate size) will reduce the tensor size, potentially leading to low MatMul efficiency. Our optimal model solves these problems by making the number of experts (256) as an exponent of 2, enables the implementation of wide-range TP(8) $\times$ EP(4) partitioning, including communication-efficient strategies such as TP-extend-EP and hierarchical EP All-to-All in Ascend Networks.

\minisection{Hidden Size vs.\ NPU Operator Shape} When considering computational efficiency on Ascend NPUs with DaVinci Architecture~\cite{DaVinci}, we aim to pursue highest utilization of MatMul operator on Cube Unit in its AI core. The Cube Unit accepts two 16 by 16 matrices for MatMul in each operation. Selecting shape as multiples of 256 for all the weights, input shapes, and output shapes ensures the size of matrices in computations to be split evenly into 16 by 16 pieces, even after possible parallelisms. Such shapes ensure maximum kernel utilization on NPU kernels during computing. Besides, by taking the computing power and communication bandwidth of Ascend NPUs into our consideration, the tensor shapes in the optimal model achieves more balanced computing and communication time, enables high degree of overlapped streams for improved throughput.

\minisection{Layer Number vs.\ Pipeline Balancing}
To alleviate on-device memory pressure, PP and VPP is usually applied for large-scale models. When separating the whole model into multiple PP chunks, load balancing of all chunk layers including possible Multi-Token Prediction (MTP) layers~\cite{deepseekv3} should be considered for reduced bubbles. MTP layers can add up to the total number of layers of the chosen model, from 61 to 64, making PP and VPP flexible for various combinations.

In summary, this section presents an analysis of the MoE architecture optimized for Ascend NPUs through simulations. While our simulation tool conducts an effective architecture search, the interpretation of how optimal model architectures work on Ascend is equally critical as a guide for general model design. The resulting model configuration is used for the system study in the following sections.

\section{MoE Training Analysis}
In this section, we dive into key analyses and decisions for training Mixture of Experts (MoE) models, highlighting aspects of unique challenges compared to training conventional dense models. Effective training of large-scale MoE models requires careful consideration of several factors that significantly impact stability, efficiency, and final model performance. Top-K routing~\cite{shazeer2017outrageously} is now widely employed in MoE models due to its simplicity and effectiveness. However, a major limitation of this unrestricted routing mechanism is its potential to create imbalances in expert utilization, commonly referred to as expert load imbalance. This imbalance poses significant challenges. (1) \emph{Training instability and routing collapse}: Without regularization, the routing network might learn to heavily favor a small subset of experts for most tokens. This can lead to these experts becoming overly specialized or saturated, while others remain underutilized. In extreme cases, this can result in routing collapse~\cite{shazeer2017outrageously}, where only a few experts effectively participate in the computation, diminishing the benefits of the MoE architecture. (2) \emph{Computational inefficiency}: MoE models often leverage expert parallelism, distributing experts across different devices. An uneven workload means some devices might be processing significantly more tokens than others. This creates computational bottlenecks, as the overall training step time or inference latency is dictated by the slowest device.

To resolve these issues, two primary strategies are often considered. Auxiliary losses are proposed to implicitly control the degree of expert balance through regularization, while the drop-and-pad~\cite{lepikhin2020gshard,fedus2022switch} strategy directly and explicitly drops tokens that exceed an expert's predefined capacity, primarily aiming to improve training throughput by maintaining a fixed computation graph. In Section~\ref{sec:load_balancing_loss}, we conduct an in-depth analysis of the load balancing loss in MoE training and propose an EP-Group auxiliary loss, which strikes a better trade-off between model performance and training efficiency. Subsequently, in Section~\ref{sec:drop_vs_dropless}, we investigate the drop-and-pad strategy. We find that while it can speed up training, it often significantly impacts model performance, especially for larger MoE models. Therefore, we adopt a dropless routing approach for \modelname. To explicitly maintain load balance without dropping tokens, we will introduce an alternative strategy involving expert placement in Section~\ref{section: Device-level Load Balance}.

\subsection{Load Balancing Loss}
\label{sec:load_balancing_loss}

To mitigate expert load imbalance, various auxiliary loss functions have been introduced. These losses are typically added to the main task loss during training to encourage a more uniform distribution of tokens across experts. Common strategies compute this auxiliary loss at different granularities, such as the sequence level~\cite{deepseekv3}, micro-batch level~\cite{lepikhin2020gshard,fedus2022switch}, and global-batch level~\cite{liu2024grin,qiu2025demonsdetailimplementingload}.
Considering a sequence with $T$ tokens, the sequence-level auxiliary loss is often defined as:
\begin{equation*}
    \ell_{\text{sequence-level}} = \alpha \sum_{i = 1}^N f_i p_i,
    \label{eq:sequence_aux_loss}
\end{equation*}
where $N$ is the total number of experts and the hyperparameter $\alpha$ controls the strength of the auxiliary loss. Here, $f_i$ represents the fraction of tokens (expert selection frequency) within the sequence routed to expert $i$, and $p_i$ is the average gating score assigned to expert $i$ over the sequence:
\begin{equation}
f_i  = \frac{N}{K T} \sum_{t \in T} \mathbbm{1} \left \{ \text{Token $t$ selects Expert $i$}\right \}, \\ \quad 
p_i  = \frac{1}{T} \sum_{t \in T} s_{i, t},
\label{selection}
\end{equation}
where $K$ is the number of experts activated per token, $\mathbbm{1}\{\cdot\}$ is the indicator function, and $s_{i, t}$ denotes the gating score of expert $i$ for token $t$. There are also other variants of auxiliary loss designed at the micro-batch level. For example, DeepSeekMoE~\cite{dai2024deepseekmoe} comes up with the device-level balance loss to ensure balanced computation across different devices. While the micro-batch level auxiliary loss is a default implementation in popular training frameworks like Megatron-LM~\cite{shoeybi2020megatronlm}, its effectiveness can be context-dependent. When training very large MoE models, memory constraints often necessitate using a very small micro-batch size per device (\eg, micro-batch size of 1). In such scenarios, the micro-batch level loss becomes nearly identical to the sequence-level loss, imposing a very strong regularization constraint. This strong regularization might negatively impact model performance by forcing the router to distribute tokens uniformly across experts. 

\begin{table*}[t]
\centering
\caption{Comparison of auxiliary loss strategies. Balance BSZ indicates the number of tokens considered when calculating the expert selection frequency. $T$ denotes the sequence length. Note that the size of the data-parallel process group $DP \ge EP$, and we only consider the case without gradient accumulation here.} 
\label{tab:aux_loss_comparison}
\begin{tabular}{ccc}
\toprule
Auxiliary Loss Level & Balance BSZ & Regularization Strength \\
\midrule
Sequence & $T$ & Strongest (local) \\
Micro-Batch  &  $\texttt{micro\_batch\_size} \times T$ & Strong (local) \\
EP-Group  & $EP \times \texttt{micro\_batch\_size} \times T$ & Medium  \\
DP-Group  & $DP \times \texttt{micro\_batch\_size} \times T$ & Weakest (global) \\
\bottomrule
\end{tabular}
\end{table*}

\begin{figure*}[b]
    \centering
    \begin{minipage}[b]{0.49\textwidth}
        \centering
        \includegraphics[width=\textwidth]{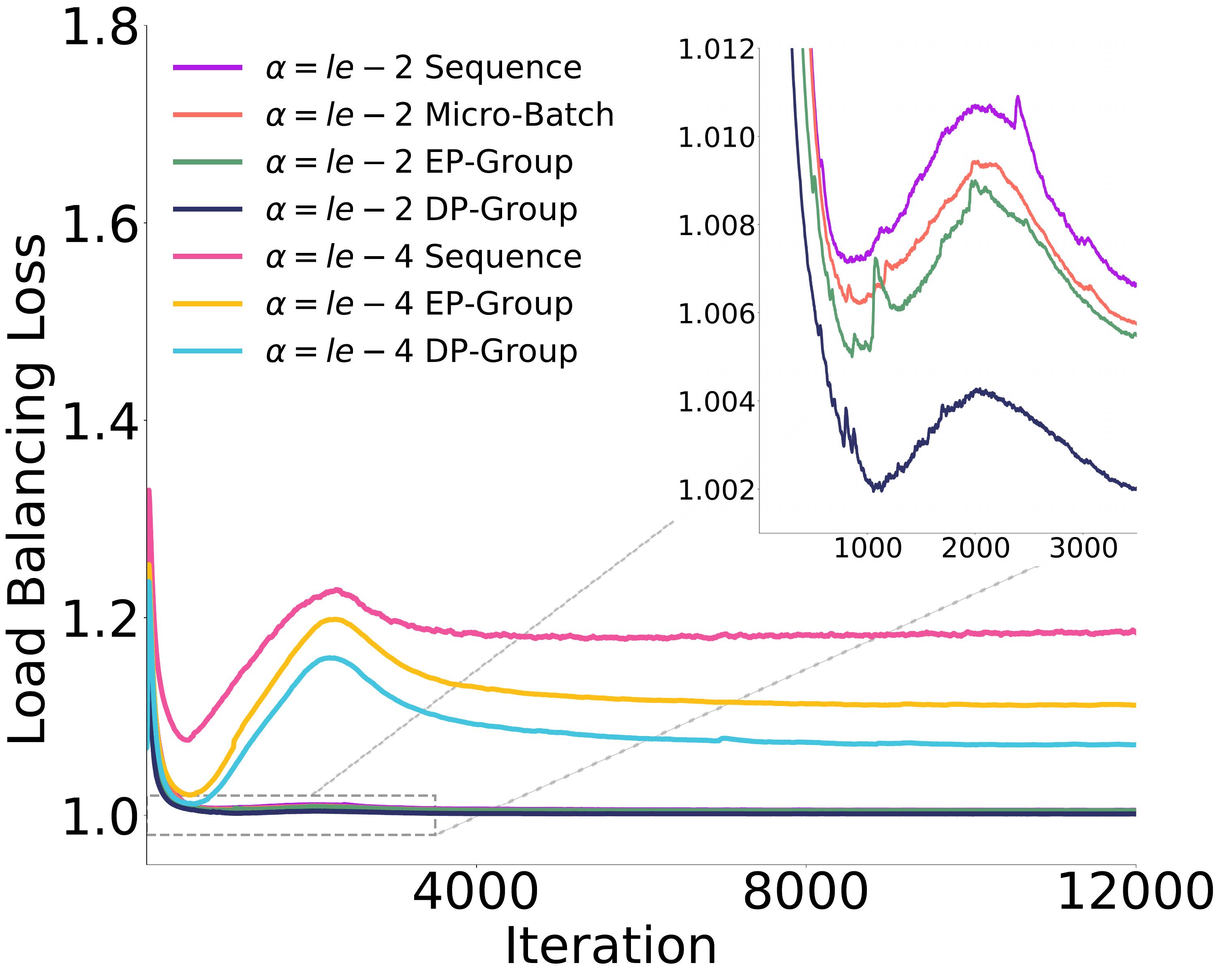}
    \end{minipage}
    \hfill
    \begin{minipage}[b]{0.49\textwidth}
        \centering
        \includegraphics[width=\textwidth]{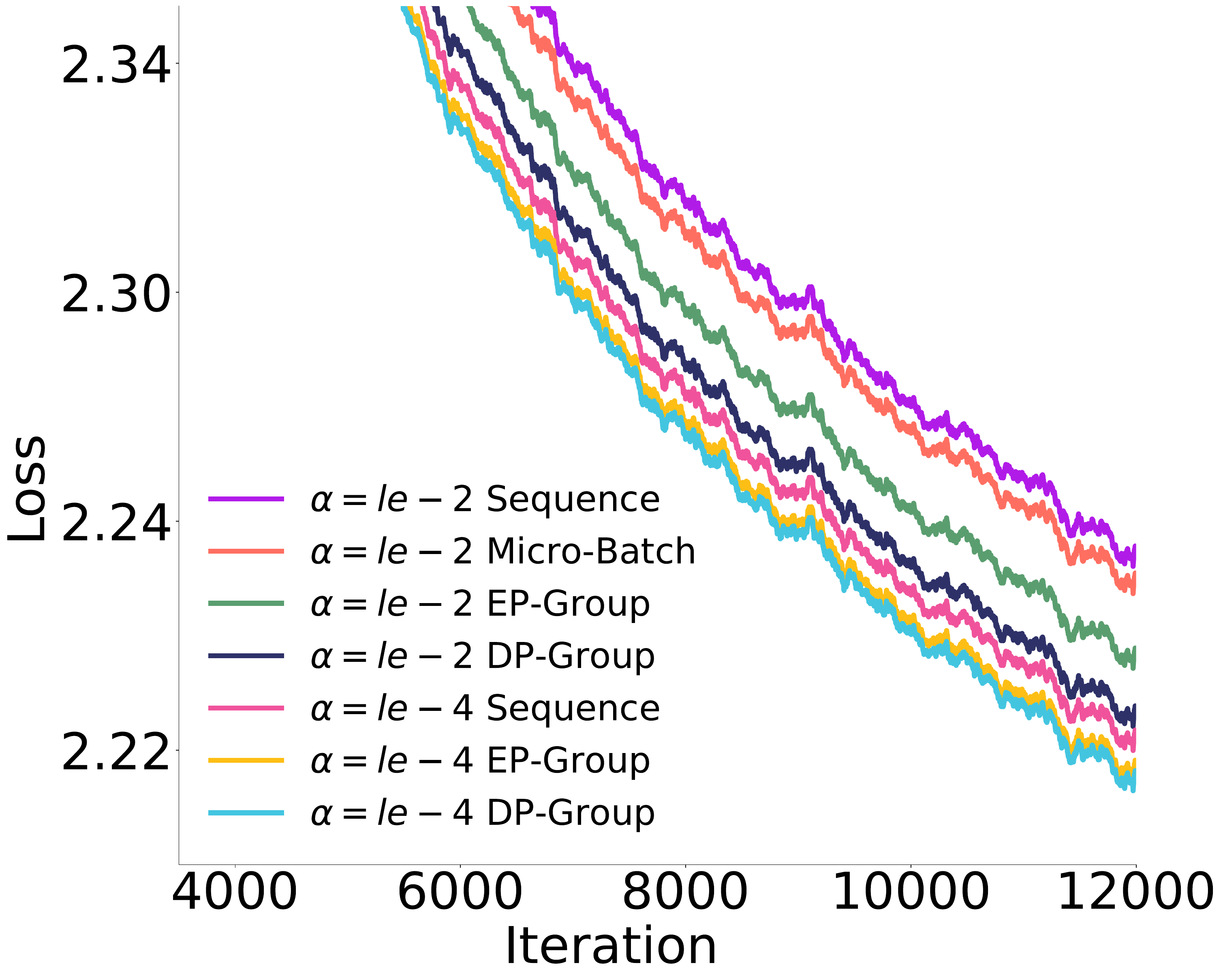}
    \end{minipage}
    \caption{Comparison of the load balancing auxiliary loss and cross-entropy loss during training across different auxiliary loss strategies and potentially varying regularization strengths, \ie, $\alpha \in \{\text{1e-2}, \text{1e-4}\}$. Clearly, setting $\alpha$ to 1e-2 more effectively controls the expert load, with the auxiliary loss remaining close to its minimum value (around 1.0). However, the right plot shows that a stronger load constraint leads to a higher pretraining loss. On the other hand, when changing the scope of tokens used to calculate the expert selection frequency, it is noticeable that different levels of auxiliary loss exhibit varying optimization difficulties. Specifically, the Sequence level is the hardest to optimize, while the DP-Group level is the easiest.}
    \label{load_balancing_loss_lm_loss}
\end{figure*}

Conversely, computing the auxiliary loss at the global-batch level provides a much weaker, coarser-grained constraint. While promoting overall balance across a large number of tokens, it might fail to prevent severe local imbalances within specific sequences or micro-batches. Such local imbalances could still lead to inefficiencies or localized training instability. To balance these trade-offs, hybrid approaches have emerged. For instance, DeepSeek-V3~\cite{deepseekv3} employs a global expert bias mechanism to manage long-term load balance while simultaneously using a weak sequence-level auxiliary loss (with a small $\alpha = \text{1e-4}$) to provide gentle local regularization. Following \citep{qiu2025demonsdetailimplementingload}, we use the term \emph{Balance BSZ} (balance batch size) to indicate the number of tokens considered when calculating the expert selection frequency $f_i$ in Eq.~\eqref{selection} for the auxiliary loss. Table~\ref{tab:aux_loss_comparison} provides a detailed comparison of different auxiliary loss strategies, outlining their respective Balance BSZ and the corresponding strength of the regularization imposed on the router network.

\minisection{EP-Group Auxiliary Loss} In our approach, we propose computing the load balance auxiliary loss at the Expert Parallelism (EP) group level. Specifically, the token counts for calculating $f_i$ are aggregated via an \texttt{all-reduce} operation only within the respective expert parallel process group for each expert. This approach strikes a balance between purely local balancing (like sequence-level) and purely global balancing (like global-batch or DP-group level). In our experiments, we compare the sequence-level, micro-batch level, DP-group level, and our proposed EP-group level auxiliary losses. For each strategy, we vary the auxiliary loss coefficient $\alpha$ between 1e-4 and 1e-2 for the baseline 20B parameter MoE model in Section~\ref{sec:moe_design}.

The comparative results are illustrated in Figure~\ref{load_balancing_loss_lm_loss}. From the figure, it can be observed that the auxiliary loss computed over the largest scope, \ie, DP-Group level, which has the largest Balance BSZ, tends to yield the lowest main training loss, suggesting potentially better final model performance when fully converged. However, due to the broader communication scope required for the \texttt{all-reduce} operation, its training throughput might be notably lower compared to the EP-Group auxiliary loss. Furthermore, when a weaker balancing strength $\alpha = \text{1e-4}$ is applied, the main training losses achieved using either the EP-Group or the DP-Group level balancing become nearly identical. Therefore, considering this interplay, the EP-Group auxiliary loss appears to offer a favorable compromise, potentially representing a better choice for balancing the trade-off between optimizing model performance and maintaining high training efficiency.

\subsection{Drop-and-Pad vs Dropless}
\label{sec:drop_vs_dropless}

In MoE training, managing the non-uniform distribution of tokens across experts represents a significant challenge. This imbalance leads to varied expert loads, impacting training efficiency and increasing the likelihood of Out-of-Memory (OOM) errors. Gshard~\cite{lepikhin2020gshard} and Switch Transformers~\cite{fedus2022switch} mitigate this issue by introducing expert capacity mechanism, which sets an upper bound on the number of tokens a single expert can process within a layer. Tokens routed to an expert that exceed this predefined capacity are not processed in that layer but are typically forwarded through a residual connection. This capacity-limiting strategy is effective in improving training speed by bounding the peak computational load on individual experts. Conversely, dropless training architectures, such as MegaBlocks~\cite{gale2023megablocks} and Deepseek-V3~\cite{deepseekv3}, process all tokens without discarding.

\begin{figure}[b]
    \centering
    \includegraphics[width=0.5\textwidth]{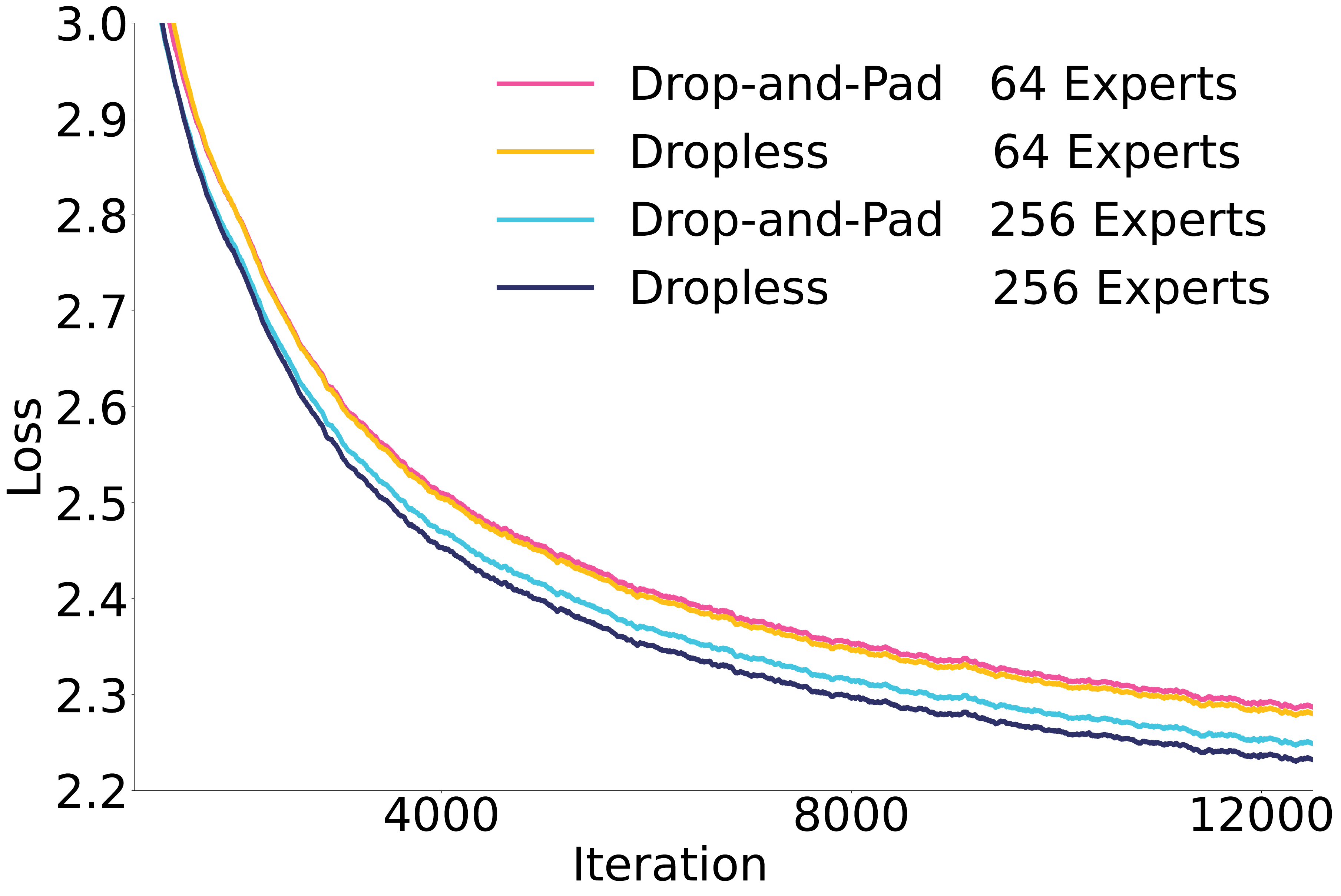} 
    \caption{Training loss curves for the 20B MoE models trained with drop-and-pad and dropless methods. Throughout training, the dropless method consistently exhibits lower loss. Specifically, with 64 experts, the dropless method yields approximately 0.008 lower loss, and this difference increases to approximately 0.017 when 256 experts are utilized.
    This indicates that the performance degradation (higher training loss) introduced by the drop-and-pad method becomes more significant as the number of experts increases.}
    \label{fig:dropless_vs_drop_loss}
\end{figure}

\begin{figure}[t]
    \centering
    \includegraphics[width=0.5\textwidth]{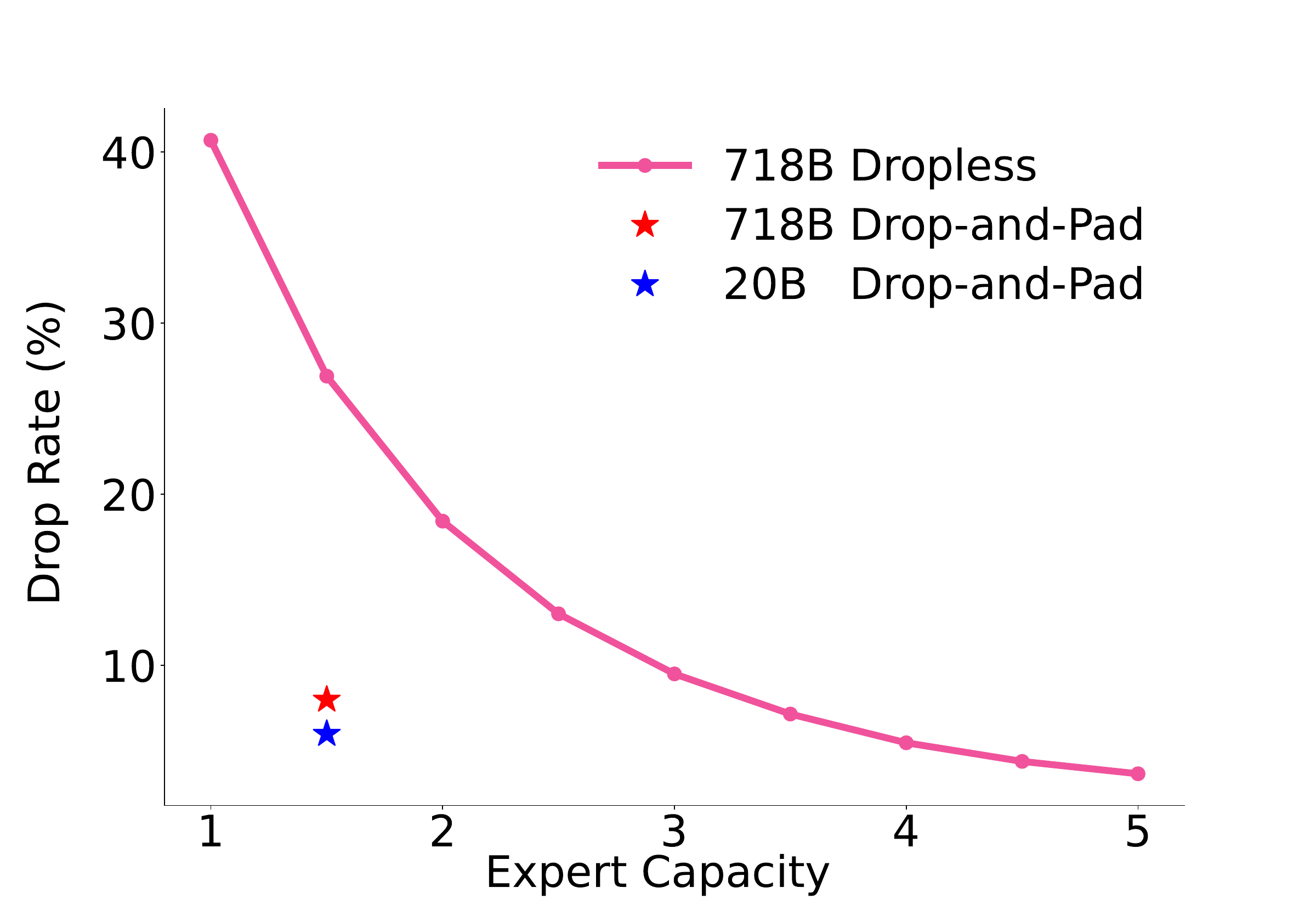} 
    \caption{Analysis of token drop rates. During drop-and-pad training with the expert capacity factor set to 1.5, larger models intrinsically drop more tokens (\ie, 8\% for a 718B MoE vs 6\% for the 20B MoE baseline), indicating potentially higher information loss and performance degradation. Separately, for \modelname{}~trained using the dropless strategy, we simulate the effect of imposing varying expert capacity limits during inference. As shown, the resulting hypothetical drop rate naturally decreases as the allowed inference capacity increases. Results are evaluated on a random 0.5\% subset of the C4~\cite{2019t5} validation data.}
    \label{Pangu Ultra MOE_drop_rate}
\end{figure}

Using the same 20B MoE setting, we conduct an experiment to compare the performance of the dropless training strategy and the drop-and-pad training strategy. In terms of experimental setup, all model parameters were kept consistent with the dropless scheme, with the exception of the MoE routing strategy. For the drop-and-pad strategy, we set the expert capacity factor to 1.5. Figure~\ref{fig:dropless_vs_drop_loss} presents the training loss curves with the two dropping strategies for the 20B MoE baseline model, where we observed significant performance degradation with the drop-and-pad strategy. This finding naturally raises the question of scalability: \emph{does this phenomenon persist at larger model sizes?} As shown in Figure~\ref{Pangu Ultra MOE_drop_rate}, for the 20B MoE model trained with the drop-and-pad strategy and an expert capacity of 1.5, the drop rate is approximately 6$\%$ (marked by a blue star). However, for the 718B counterpart, with the same expert capacity and drop-and-pad strategy, the drop rate increases to around 8$\%$ (marked by a red star). This suggests that, under the drop-and-pad strategy, larger models experience higher drop rates, leading to a more pronounced performance loss.

Therefore, for larger-scale models, the dropless strategy is more favorable, as it better preserves model accuracy without the performance degradation associated with token dropping. Prioritizing high model performance, our system-level efforts in Section~\ref{section: Training System Optimization} would focus on improving the training efficiency of the dropless strategy.

\section{Training System Optimization}
\label{section: Training System Optimization}
The training of the \modelname~is powered by MindSpeed~\cite{mindspeed} platform and Megatron~\cite{shoeybi2019megatron} framework. To further improve training efficiency and stability, we have implemented a suite of advanced engineering optimizations across four essential areas: Parallelization Strategies, Communication Optimization, Memory Optimization, Expert Load Balancing Optimization. These optimizations collectively address computational, communication, and memory bottlenecks, enhancing system performance at a remarkable scale. 

\subsection{Parallelism Optimization}
\label{section: Parallelization Strategy Optimization}
The growing complexity and scale of LLMs demand innovative approaches to enhance computational efficiency without compromising performance. A critical aspect of this pursuit lies in optimizing parallelism strategies, which enables efficient distribution of workloads across heterogeneous hardware systems. This section explores our methodology for designing and implementing tailored parallelism techniques to unlock scalable, cost-effective deployment on Ascend NPUs while preserving the model’s expressive capabilities.

\minisection{Parallelism Strategy Design} We leverage the end-to-end modeling and simulation platform mentioned in Section~\ref{section: Simulation} to systematically explore the design space and automatically discover the optimal multi-dimensional parallelism strategies. To ensure better load balancing across devices, we initially set MBS = 2. As identified from simulation, the optimal parallelism strategy on 6K Ascend NPUs is: TP = 8, PP = 16, VPP = 2, and EP = 4. Tensor Parallelism (TP) partitions the model to minimize per-device memory usage while leveraging high-speed intra-node interconnects. By integrating communication over computation (CoC)~\cite{coc2025}, communication exposure in TP is effectively mitigated. Our evaluation revealed that Tiling = 4 achieves approximately 4× reduction in communication exposure, with negligible impact on computational throughput. We also utilized Pipeline Parallelism (PP) to decrease the number of layers per device. We implemented pipeline interleaving to effectively mitigate idle periods caused by pipeline bubbles.

For Expert Parallelism (EP), we adopt a TP-extended EP strategy, where the expert dimension is not partitioned along TP to avoid inefficient small tensor shapes. Instead, experts are distributed at the expert granularity. The EP degree is determined through a joint optimization of memory usage, communication overhead, and token routing imbalance, with EP = 4 yielding the optimal trade-off. Given TP = 8, the total number of expert groups is 32 (TP $\times$ EP). To minimize EP-related communication, particularly across machines, we implement hierarchical schemes such as group-wise All-to-All \cite{li2024locmoe} to minimize expert communication costs. More detailed implementation is introduced in Section~\ref{section: Hierarchical All-to-All Communication}.

\minisection{Load-Balanced Virtual Pipeline Parallelism Design}
The \modelname{} training framework achieves a pipeline bubble ratio of 10.49\%  through Virtual Pipeline Parallelism (VPP), representing a 1.8× improvement over traditional pipeline parallelism, which exhibits a bubble ratio of 18.98\%. This reduction in pipeline bubbles is primarily attributed to the balanced distribution of computational loads across pipeline stages. However, the Multi-Token Prediction (MTP) layer poses significant load balancing challenges due to its aggregated computational demands. Specifically, the combined workload of the MTP layer, output head, and cross-entropy loss calculation is approximately 2.5$\times$ the execution time of a standard MoE layer, surpassing the stage capacity optimized for 2$\times$ MoE operations. This 25\% overhead would impose severe pipeline backpressure if concentrated within a single stage, causing waiting bubbles to propagate upstream across the pipeline.

\begin{figure*}[t]
    \centering
    \includegraphics[width=0.8\textwidth]{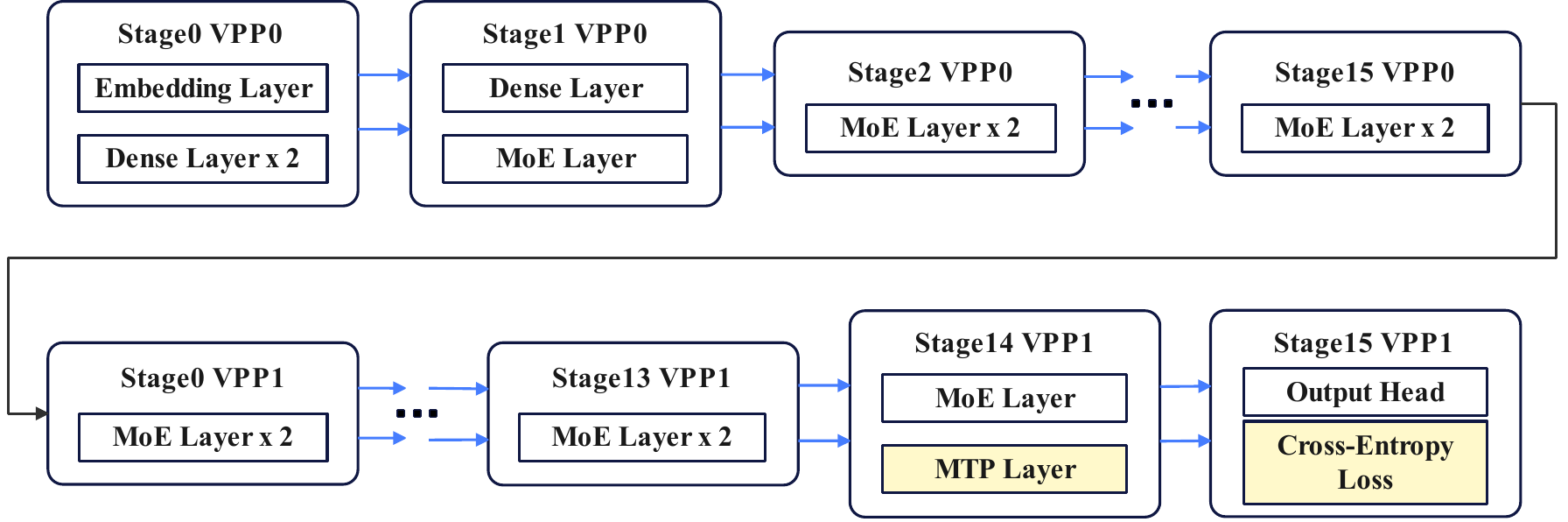}
    \caption{Virtual pipeline stage partitioning matrix (2-row VPP Stages × 16-column PP Stages). The MTP layer is assigned to Stage14 of VPP1, and the cross-entropy loss function is placed in Stage15 of VPP1.}
    \label{fig:vpp_flowchart}
\end{figure*}

To resolve this, we strategically distribute the computational components across multiple stages: \texttt{Stage14\_VPP1} is assigned the MTP body (equivalent to 1.05$\times$MoE) along with one standard MoE layer, while \texttt{Stage15\_VPP1} is responsible for processing the output head and loss calculations (equivalent to 1.5$\times$MoE). This staged configuration constrains workload overflow to a 5\% tolerance (2.05$\times$MoE vs 2.0$\times$MoE baseline) by leveraging pipeline prefetching mechanisms, thereby effectively mitigating potential performance degradation. The VPP partitioning strategy is illustrated in Figure~\ref{fig:vpp_flowchart}. Moreover, The word embeddings for both MTP tokens are entirely computed within the first pipeline stage, eliminating cross-stage gradient synchronization of word embeddings. Although this doubles the inter-stage point-to-point(P2P) communication volume (from 1$\times$ to 2$\times$ tokens), the Adaptive Pipe Overlap scheduling strategy introduced in Section~\ref{section: 1F1B Communication Overlap Design} completely masks this overhead through computation-communication overlap.

\subsection{Communication Optimization}
\label{section: Hierarchical All-to-All Communication}
As parallelism increases, communication overhead often becomes a critical bottleneck, making communication optimization crucial to effectively exploit parallelism, minimize latency, and maximize throughput across distributed computational units. We propose Hierarchical EP All-to-All Communication strategy to efficiently arrange the inter-node and intra-node communications, and Adaptive Pipe Overlap approach to effectively mask the communication overhead. 

\minisection{Hierarchical EP All-to-All Communication} In the Megatron framework, AllGather and All-to-All represent two distinct token dispatching mechanisms designed to efficiently distribute input tokens to experts. The AllGather mechanism operates by broadcasting local tokens from each NPU to all other NPUs, ensuring that each NPU retains complete global token information. However, this approach incurs significant communication overhead, particularly when scaling to a large number of NPUs. In contrast, the All-to-All mechanism significantly optimizes communication by directly routing tokens to their target NPUs based on their expert assignments, leading to superior communication efficiency by eliminating redundant data transmission, making it more suitable for large-scale training.

Despite its advantages, the All-to-All mechanism has limitations. Experts are distributed across multiple nodes when the EP size exceeds the number of NPUs in a single node, leading to inter-node communication with much lower bandwidth than intra-node bandwidth. To address this, we propose a Hierarchical Expert-Parallel Communication Mechanism which strategically restructures the communication flow to prioritize intra-node transfers in two phases. The first phase is Inter-Node AllGather Synchronization, where NPUs with the same rank across nodes first perform an AllGather to synchronize token data globally. The second phase  Intra-Node All-to-All Redistribution follows a token permutation step, where each node selects only the tokens relevant to its local experts and then conducts an optimized All-to-All exchange within the node.

This hierarchical approach converts most inter-node communication into intra-node communication at a cost of minimal inter-node redundancy. Additionally, splitting the communication operator into two stages allows for overlapping communication with computation during both forward and backward passes. Moreover, forward inter-node communication can be parallelized with backward intra-node communication, further enhancing training efficiency. Comparison of communication volumes across these dispatching mechanisms is summarized in Table~\ref{tab:Hierarchical alltoall Communication}.

\begin{table}[t] 
\centering 
\caption{Comparison of communication volumes. Hierarchical EP All-to-All Communication converts inter-node All-to-All to intra-node All-to-All. Total communication volumes are reduced. Inter-node and Intra-node communication can both be overlapped more effectively with computation.} 
\label{tab:Hierarchical alltoall Communication} 
\begin{tabular}{cccc} 
    \toprule 
    Dispatcher Type & Allgather & All-to-All & Pangu Hierarchical \\ 
    \midrule 
    \multirow{2}{*}{Inter-node} & Seq $\times$ TP $\times$ EP & Seq $\times$ TopK & Seq $\times$ (EP-1) \\ 
     & (Allgather) & (All-to-All) & (Allgather) \\
    \multirow{2}{*}{Intra-node} & \multirow{2}{*}{-} & \multirow{2}{*}{-} & Seq $\times$ TopK \\
     & & & (All-to-All) \\
    \bottomrule 
\end{tabular} 
\end{table}

\begin{figure*}[ht!]
    \centering
    \includegraphics[width=\textwidth]{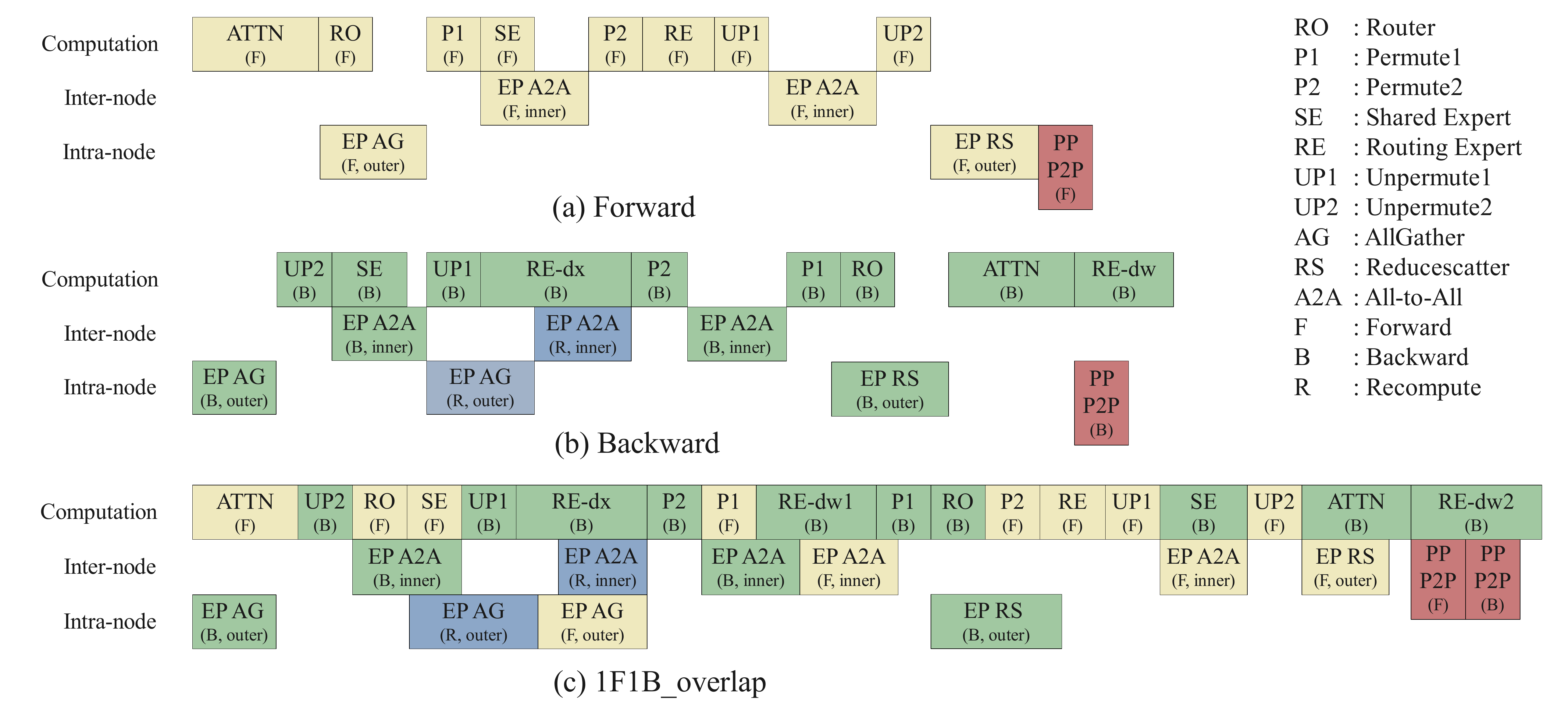}
    \caption{MoE operator streams. Subfigure (a) shows the forward pass. Subfigure (b) shows the backward pass. Subfigure (c) illustrates the Adaptive Pipe Overlap stream. The color-coded operators represent: yellow for forward, green for backward, blue for recomputation, and brown for P2P communication.}
    \label{fig:1F1B_overlap}
\end{figure*}

\minisection{Adaptive Pipe Overlap Mechanism}
\label{section: 1F1B Communication Overlap Design}
Despite the implementation of Hierarchical Expert-Parallel Communication, the expert-parallel communication volume remains substantial. Traditional self-overlap strategies are insufficient to fully mask this communication overhead. The DeepSeekV3 framework propose a dual-pipe forward-backward overlapping approach~\cite{deepseekv3}, which leverages computations from different microbatches to overlap communication, reducing bubble time by half. However, the dual-pipe pipeline doubles static memory footprint and still suffers from high bubble rates. To address this, we propose a VPP-based Adaptive Pipe Overlap Strategy(1F1B\_overlap), which innovatively exploits the independence between micro-batches to mask backward communications with forward computations (and vice versa). Additionally, we incorporated optimizations for overlapping permute recomputation communications, overlapping PP communications, and alleviating host-bound bottlenecks, as illustrated in Figure~\ref{fig:1F1B_overlap}.

The core design features of Adaptive Pipe Overlap include Hierarchical Communication Overlap, Mitigating Host-Bound Bottlenecks and Decoupling Backward of Routing Experts, summarized below.

The outer EP involves two AllGather and two ReduceScatter operations per 1F1B\_overlap iteration, each with significant latency. These are overlapped with the forward and backward computation of attention and routing experts, respectively. The inner EP involves four AlltoAllv operations per 1F1B\_overlap iteration, each with shorter latency. These are overlapped with computations of the router and shared experts. By leveraging independent communication links, the inner EP and outer EP mutually overlap each other. The TP communication achieves self-overlapping through fused operators. Utilizing distinct communication links, it also mutually overlaps with the outer EP.

\begin{figure*}[ht!]
    \centering
    \includegraphics[width=0.8\textwidth]{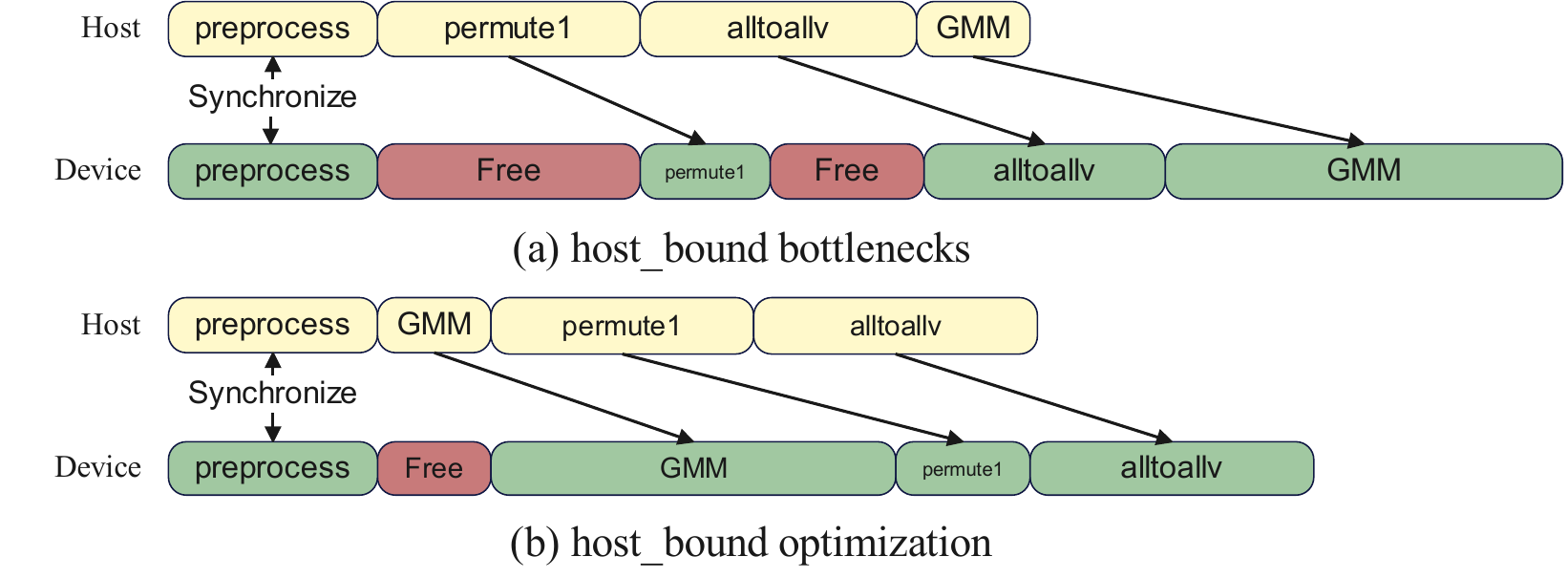}
    \caption{Comparison between before host\_bound optimization and after. Compared with permute1, GMM incurs less host runtime but longer computation time on device. Scheduling GMM ahead of Permute1 mitigates the host-bound impact.}
    \label{fig:host_bound optimization}
\end{figure*}

Host-bound bottlenecks stem from synchronization during preprocess in the MoE module. Preprocess constructs input splits and output splits for AlltoAllv communication, which are transferred from device to host (D2H), introducing synchronization. The immediate scheduling of permutate1 and AlltoAllv communication after preprocess causes significant free on device, as illustrated in Figure~\ref{fig:host_bound optimization}~(a). To address this problem, we decoupled preprocess from permutate1 and immediately schedule the computationally intensive GMM operator after preprocess, thereby minimizing synchronization-induced host-bound latency as demonstrated in Figure~\ref{fig:host_bound optimization}~(b).

In the Adaptive Pipe Overlap schedule, no operator depends on the dw (weight gradient) of routing experts. Thus, the dx (input gradient) and dw of routing experts can be decoupled, allowing dw to be flexibly repositioned to overlap with communication. Specifically, routing experts involve gradient computations for two weights (w1 and w2). The decoupled dw1 overlaps with intra-layer AlltoAllv communication. The decoupled dw2 from each layer in this VPP stage collectively overlaps with PP communication.

Experimental results reveal that Adaptive Pipe Overlap has achieved a 95\% communication overlap rate across TP, EP, and PP, where specialized forward-backward overlapping for MoE modules ensures robust load balancing. While preserving their native communication paradigms, non-optimized components (e.g., dense/output layers) maintain lower latency than the MoE layers.

\subsection{Memory Optimization}
\label{section: Memory Optimizations}
Memory constraints significantly restrict the flexibility of model configurations and parallelism strategies, thereby limiting training performance. We introduced two core techniques to optimize memory utilization on Ascend NPUs: fine-grained recomputation and tensor swapping. These methods focus on optimizing activation memory in specific modules (such as MLA, permute, and activation) rather than recomputing entire layers. This targeted strategy enables greater stability, scalability, and cost-effectiveness throughout of the LLM training process.

\minisection{Fine-Grained Recomputation}
While Megatron's existing transformer layer recomputation and self-attention recomputation mechanisms~\cite{megatron-github2025url} provide foundational memory optimization capabilities, the inherent trade-off between memory savings and added computational overhead motivates our pursuit of finer-grained recomputation strategies~\cite{chen2016training,peng2020capuchin}. We implemented a targeted recomputation approach through three specialized modules: MLA, permute and activation. This achieves superior efficiency over layer or self-attention recomputation, by eliminating unnecessary recomputation.
 
Multi-head latent attention (MLA)~\cite{liu2024deepseek,deepseekv3} is an advanced attention mechanism designed to shrink the KV cache by compressing keys and values into a smaller, low-rank matrix. Our proposed MLA QKV recomputation releases the activation memory for queries, keys and values, with recomputation of QKV starting from up-projection, as down-projection process involves low activation memory but requires long execution time. We also propose an alternative KV-only recomputation strategy, leveraging that the queries computation can be entirely decoupled from keys/values computation. Although this version can only save the activation memory of keys and values, it eliminates the queries recomputation overhead. When memory is extremely constrained, MLA QKV recomputation is prioritized. When there is enough room to accomodate queries activation memory, KV-only version outperforms QKV version with less recomputation time.

The permute operation incurs the largest activation memory overhead. Recomputing the permute operation can significantly alleviate memory pressure, making it the preferred choice when memory usage is constrained. The SwigLU activation between group expert linear layers is the most efficient recomputation strategy, offering the best memory-to-time trade-off. 

A significant advantage of these fine-grained recomputation strategies is their ability to be flexibly scheduled and overlapped with communication during 1F1B\_overlap iteration. This enables the activation memory savings to compensate for the additional recomputation time, effectively mitigating the associated overhead and minimizing the negative impact on overall performance.

\minisection{Tensor Swapping}
Tensor swapping~\cite{le2019tflmslargemodelsupport,meng2017training,huang2020swapadvisor} optimizes device memory by temporarily offloading unused parameters, gradients, or activations to host memory during the forward pass and reloading them when required for backward computations. To improve training performance, activations with high recomputation cost are swapped rather than recomputed, saving the runtime overhead associated with recomputation. 

The computation of probs during backpropagration requires tokens from forward unpermutation, which provides substantial memory savings when being swapped. The offloading process begins concurrently with Token Unpermutation computation and concludes across micro-batch, finishing before the next Token Unpermutation computation. During backward pass, token activations required in current micro-batch are prefetched simultaneously with tokens gradient computation at Token Unpermutation in the previous micro-batch.

With a variety of memory-saving strategies available, it's possible to effectively apply different combinations to explore the memory limits of NPUs while minimizing the additional runtime overhead. Our training strategy utilizes a combination of MLA KV-only recomputation, permute recomputation, activation recomputation, and prob swapping, accelerating the training process by replacing the recomputation of entire layers.

\subsection{Load Balancing Optimization}
\label{section: Device-level Load Balance}
\begin{figure*}[ht!]
    \centering
    \includegraphics[width=\textwidth]{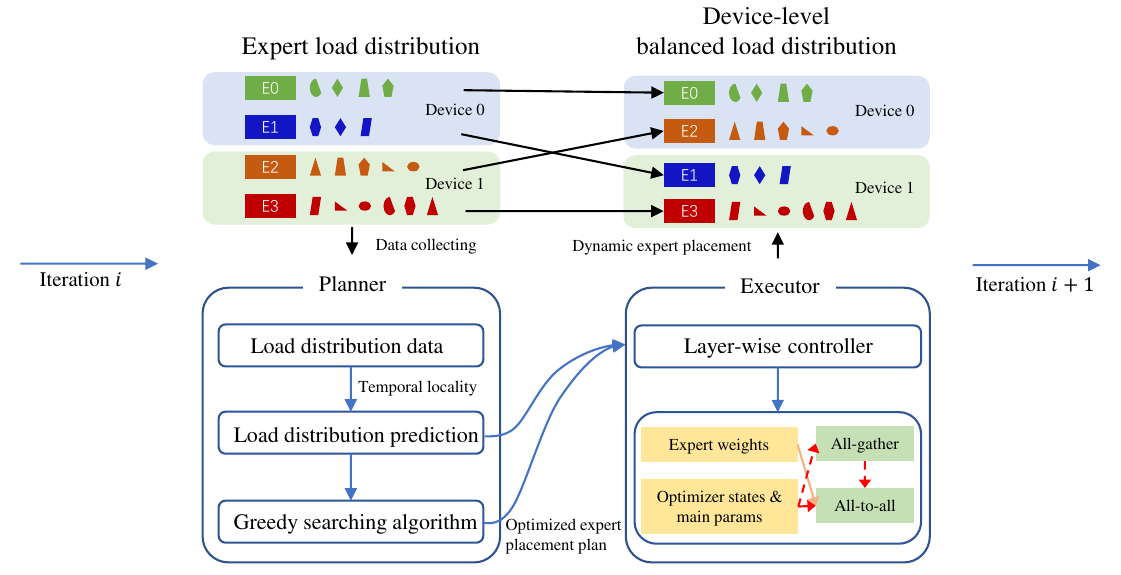}
    \caption{Overview of dynamic device-level load balancing mechanism. The planner generates and periodically updates the expert placement strategy based on load distribution predictions. While the executor performs dynamic layer-wise expert placement and manages the swapping of expert parameters and optimizer states.}
    \label{fig:load_balance}
\end{figure*}

For MoE model training on Ascend NPUs, even distribution of tokens per device (device-level load balance) is critical for the performance of above optimization strategies and training stability. Auxiliary losses implicitly balance expert loads but may leave residual imbalance~\cite{li2024locmoe,wei2024skywork}. To resolve this problem, we propose an explicit dynamic device-level load balancing mechanism with a planner and an executor as shown in Figure~\ref{fig:load_balance}, which dynamically adjusts expert placement across devices, achieving a relative 10\% improvement in MFU. Our method incurs smaller memory overhead compared to FlexMoE~\cite{nie2023flexmoe}, thus avoiding additional memory pressure. Additionally, layer-wise dynamic expert placement reduces computational costs relative to SmartMoE's fixed-frequency placement~\cite{smartMoE}.

\minisection{Planner} 
To achieve device-level load balancing, the planner generates an expert placement strategy that is periodically switched at a specified interval based on historical load distribution data. A sliding window average method is employed to forecast load distribution~\cite{cong2024prediction}, followed by a light-weight greedy algorithm to find a  plan that balances the predicted load distribution. To ensure global consistency, the predicted load distribution is synchronized across all expert data-parallel groups. Predictions and expert placement searching  are guaranteed to be effective by the temporal locality inherent in load distribution patterns. For MoE models with fine-grained experts, our approach could reduce device-level load imbalance by 80\%-90\%.

\minisection{Executor}
The executor performs layer-wise dynamic expert placement before the next forward pass, fully decoupled from other optimizations. Expert parameters and optimizer states are swapped through efficient All-to-All communication. In mixed-precision training with distributed optimizers~\cite{zhai2023smartmoe}, main parameters and optimizer states  in FP32 are gathered before swapping. To minimize communication overhead caused by frequent expert placements, a controller triggers layer-specific placement only when load imbalance increases (measured by coefficient of variation). This achieves layer-wise, fine-grained dynamic expert placement. As load distribution stabilizes during supervised fine-tuning, the expert placement is executed only once for maximum efficiency.

\subsection{Other Optimizations}
\label{section: Optimizations Related to the Host}
Beyond the aforementioned optimization strategies, we propose additional techniques tailored for Ascend NPUs, including host-bound optimization, computation offloading and data sharing methods, and kernel fusion. These innovations further accelerate the training process and enhance overall system efficiency.

\minisection{Host-Bound Optimizations}
During large-scale cluster training, host-bound issues pose a significant bottleneck to training efficiency. Due to its stochastic nature, different nodes may encounter host-bound bottlenecks at varying times, triggering frequent synchronization waits across the cluster and degrading overall performance. To mitigate this, synchronization-heavy operators are reduced to reduce unnecessary synchronization points. Additionally, we implement CPU Core Binding to optimize CPU affinity. On the Ascend platform, the operator dispatch on the host side and the operator execution on the device side can be carried out asynchronously by enabling the TaskQueue. When the host side can efficiently dispatch a large number of tasks to ensure the continuous operation of the device side without idle time, the Ascend NPUs can achieve their peak performance.

\minisection{Computation Offloading and Data Sharing}
In current training pipelines, some operations suffer from low efficiency on NPUs due to frequent data synchronization and complex control flow. Meanwhile, tensor transfers within the TP domain poses performance bottlenecks~\cite{narayanan2021efficient}. To address these challenges, we propose optimizations focusing on computation offloading and data sharing. These strategies collectively enhances intra-node computational and data transfer efficiency.

Non-parallelizable computations with low complexity are decoupled from the main computation graph and offloaded to the CPU during the data loading phase. These computations are well-suited for offloading, as they primarily depend on the input sequences and global parameters. Combined with TP-level tensor sharing mechanisms, this enables a "generate-once, share-across-devices" strategy that effectively reduces redundant computation and NPU memory usage.

To improve data sharing efficiency within TP domains, we propose a lightweight, lock-free tensor access protocol utilizing a shared memory mechanism~\cite{mojumder2020mgpu}. This protocol employs status flags and reference counting to manages tensor life-cycle states (ready, consumed, releasable), ensuring secure and efficient access across processes with minimal control logic.

\minisection{Kernel Fusion}
Kernel fusion is a critical optimization technique in LLM training that combines a series of discrete computational operations into a single, unified kernel to minimize overhead and maximize hardware utilization. In addition to the FlashAttention and RMSNorm fused operators already implemented in the PanGu dense model~\cite{yin2025panguultrapushinglimits}, we introduce GMMAdd, Permute and Umpermute fused operators in our MoE model. The GMMAdd fused operator combines the GroupedMatMul backward computation with gradient accumulation, leveraging parallel processing and pipeline overlap to reduce scheduling overhead. Multiple granular operations are fused in the Permute and Unpermute operators to alleviate memory access time. These fused kernels minimize data movement between compute units, improve memory locality, and maximize accelerator utilization, ultimately enhancing training throughput and resource efficiency.

In summary, a variety of optimization strategies have been introduced to enhance overall training efficiency. To quantify the impact of these strategies, Table~\ref{tab:Training throughput increase} categorizes them into distinct groups and illustrates their respective relative increase in MFU. Our implemented strategies collectively improve MFU by 58.7\% over the baseline, achieving an MFU of 30.0\% on 6K Ascend NPUs. Moreover, the finer-grained overlapping techniques and new fused operators for training MoE models on Ascend NPUs are currently in development. The further improvements in MFU will be released officially in upcoming publications.

\begin{table}[t] 
\centering 
\caption{MFU Increase of 718B MoE training on 6K Ascend NPUs. Cumulative MFU improvements are evaluated for four core strategies, each showing a relative increase over the baseline.} 
\label{tab:Training throughput increase} 
\begin{tabular}{lcc} 
    \toprule 
    Strategy & MFU increase \\ 
    \midrule 
    baseline & 1\\ 
    + fine-grained recompute and swap & 15.8\%\\
    + adaptive pipe overlap & 28.6\% \\
    + host optimization & 49.2\% \\
    + fused operators & 58.7\% \\
    \bottomrule 
\end{tabular} 
\end{table}

\section{Experiments}
Testing the software and hardware systems requires a large MoE model with state-of-the-art performance. As detailed in Section 2, our model configuration comes from a systemic search in simulation. In this section, we first present the implementation details of model training. We then conduct a comprehensive evaluation comparing instruct models with state-of-the-art MOE models. Finally, to better understand large-scale MOE, we analyze the behaviors of experts systematically.

\subsection{Implementation Details}
During the construction of the training dataset, we implement strict data quality control and emphasize the diversity, complexity, and comprehensiveness of the corpus.
For long CoT samples, we introduce specialized tokens to structurally separate reasoning trajectories from final answers.
In post-training stage, instruction fine-tuning integrates multi-domain core tasks such as general question and answer, text generation, semantic categorization, code programming, mathematical and logical reasoning, and tool usage to form a multi-dimensional training space for enhanced generalization. Additionally, we set the ratio of reasoning to non-reasoning samples to 3:1, further improving reasoning ability.

\subsection{Evaluation Results}

We evaluate the \modelname~chat version across two key dimensions: general language comprehension and reasoning capabilities:(1) general language comprehension as measured by standardized benchmarks including C-Eval~\cite{huang2023c}, CLUEWSC~\cite{xu2020clue}, IF-Eval~\cite{zhou2023instruction}, MMLU~\cite{hendrycks2020measuring}, and MMLU-PRO~\cite{wang2024mmlu}; and (2) complex reasoning capabilities, demonstrated through mathematically intensive challenges (AIME2024~\cite{MAA}, AIME2025~\cite{MAA2025}, MATH500~\cite{lightman2023let}), programming evaluations (MBPP+~\cite{austin2021program}, LiveCodeBench~\cite{jain2024livecodebench}), and advanced scientific reasoning assessments (GPQA-Diamond~\cite{rein2024gpqa}).

\begin{table}[!h]
    \centering
    \footnotesize
    \setlength{\tabcolsep}{4.5pt}
    \caption{Comparison of \modelname~and other representative MoE instruct models across diverse benchmarks for evaluating general language comprehension and reasoning skills. Bold values indicate the best results in each row; asterisks ($^*$) denotes results obtained through our own testing.}
    \small
    \begin{tabular}{@{}c l |c c c c | c@{}}
    \toprule
    & \multirow{2}{*}{\centering \textbf{Benchmark}} &  \textbf{Qwen2.5} &  \textbf{MiniMax} & \textbf{DeepSeek} & \textbf{DeepSeek} & \multirow{2}{*}{\textbf{\modelname}} \\
    &  &\textbf{Plus} & \textbf{Text-01} & \textbf{V3-0324} & \textbf{R1} & \textbf{} \\
    \midrule
    & Architecture & MoE & MoE& MoE & MoE & MoE \\
    & \# Activated Params & - & 46B & 37B & 37B & 39B \\
    & \# Total Params & - & 456B & 671B & 671B & 718B \\
    \midrule
    \multirow{6}{*}{General} &C-Eval & - & - & 87.9$^*$ & \textbf{91.8} & 90.8  \\
    & CLUEWSC & -  &- & 94.3$^*$ &  92.8 & \textbf{94.8}  \\
    & C-SimpleQA & -  & 67.4 & \textbf{72.4$^*$} & 63.7 & 70.3  \\
    & IF-Eval & 86.3 & \textbf{89.1} & 85.9$^*$ & 83.3 & 84.9\\
    & MMLU & - & 88.5 & 87.4$^*$ & 90.8 & \textbf{91.5}  \\
    & MMLU-Pro & 72.5 & 75.7 & 81.2 & \textbf{84.0} & 83.5  \\
    \midrule
    \multirow{5}{*}{Reasoning} 
    & AIME2024  & -  & - & 59.4 & 79.8 & \textbf{81.3}  \\
    & AIME2025  & - & - & 39.8$^*$ & \textbf{70.0} & \textbf{70.0}  \\
    & GPQA-Diamond & - & 54.4 & 68.4 & 71.5 & \textbf{75.3} \\
    & LiveCodeBench & 51.4& - & 49.2 & \textbf{65.9} & 64.3 \\
    & MBPP+ & - & 71.7 & 77.3$^*$ & \textbf{81.2$^*$} & \textbf{81.2} \\
    & MATH500  & - & - & 94.0 & 97.3 & \textbf{97.4}  \\    
    \bottomrule
    \end{tabular}
    \label{tab:StandardBenchmarks of Post-training}
\end{table}

Table~\ref{tab:StandardBenchmarks of Post-training} presents a comparative analysis of the \modelname~instruct model alongside strong MOE instruct models, including Qwen2.5-plus~\cite{yang2024qwen2}, MiniMax-Text-01~\cite{li2025minimax}, Deepseek-V3-0324~\cite{deepseekv3} and reasoning model Deepseek-R1~\cite{deepseekai2025deepseekr1incentivizingreasoningcapability}. \modelname~demonstrates competitive performance across multiple domains. On knowledge-intensive tasks including C-Eval, MMLU, and MMLU-Pro, our model achieves comparable scores to DeepSeek-R1 while outperforming other baseline models, showing strong knowledge comprehension capabilities. \modelname~exhibits exceptional performance in reasoning tasks, achieving SOTA results on the mathematics benchmarks. In particular, \modelname~achieves 81.3\% accuracy on AIME2024. Similar competitive performance is observed in coding-related tasks and scientific reasoning tasks.

\begin{table}[!h]
    \centering
    \footnotesize
    \setlength{\tabcolsep}{4.5pt}
    \caption{Comparison of \modelname~and DeepSeek-R1 across a set of benchmarks for Industry Assessment. Bold values represent the best results in each line.}
    \small
    \begin{tabular}{l | c | c@{}}
    \toprule
    \textbf{Benchmark}  & \textbf{DeepSeek R1} &  \textbf{\modelname} \\
    \midrule
     MedQA  & 85.8 & \textbf{87.1} \\
     MedMCQA  & 78.6 & \textbf{80.8} \\
    
    \bottomrule
    \end{tabular}
    \label{tab:StandardBenchmarks of Industry Assessment}
\end{table}

\minisection{Evaluation Results of Industry Assessment}
To comprehensively assess model capabilities across different professional domains, we conducted systematic evaluations on representative downstream tasks from key industries. Specifically in the medical domain, two benchmark datasets were selected for rigorous testing: MedQA~\cite{jin2021disease}, a comprehensive collection of medical knowledge questions designed to evaluate clinical reasoning abilities, and MedMCQA~\cite{pal2022medmcqa}, a large-scale assessment tool containing authentic questions from medical licensing examinations that tests practical diagnostic competence. The evaluation results in Table~\ref{tab:StandardBenchmarks of Industry Assessment} reveals a particularly noteworthy performance of \modelname~in medical applications. Our model present strong capabilities in processing complex medical terminology, interpreting clinical scenarios, and providing diagnostically relevant answers, achieving state-of-the-art performance on both benchmarks. This medical domain superiority suggests strong potential for specialized applications in healthcare AI systems.

\subsection{MoE Analysis}
To further investigate the advantages of the MoE architecture for \modelname, we conduct a comprehensive analysis of several key properties specific to MoE, including Domain Specialization, Router Scale, and Expert Co-Activation.

\begin{figure*}[t]
    \centering
    \includegraphics[width=0.9\textwidth]{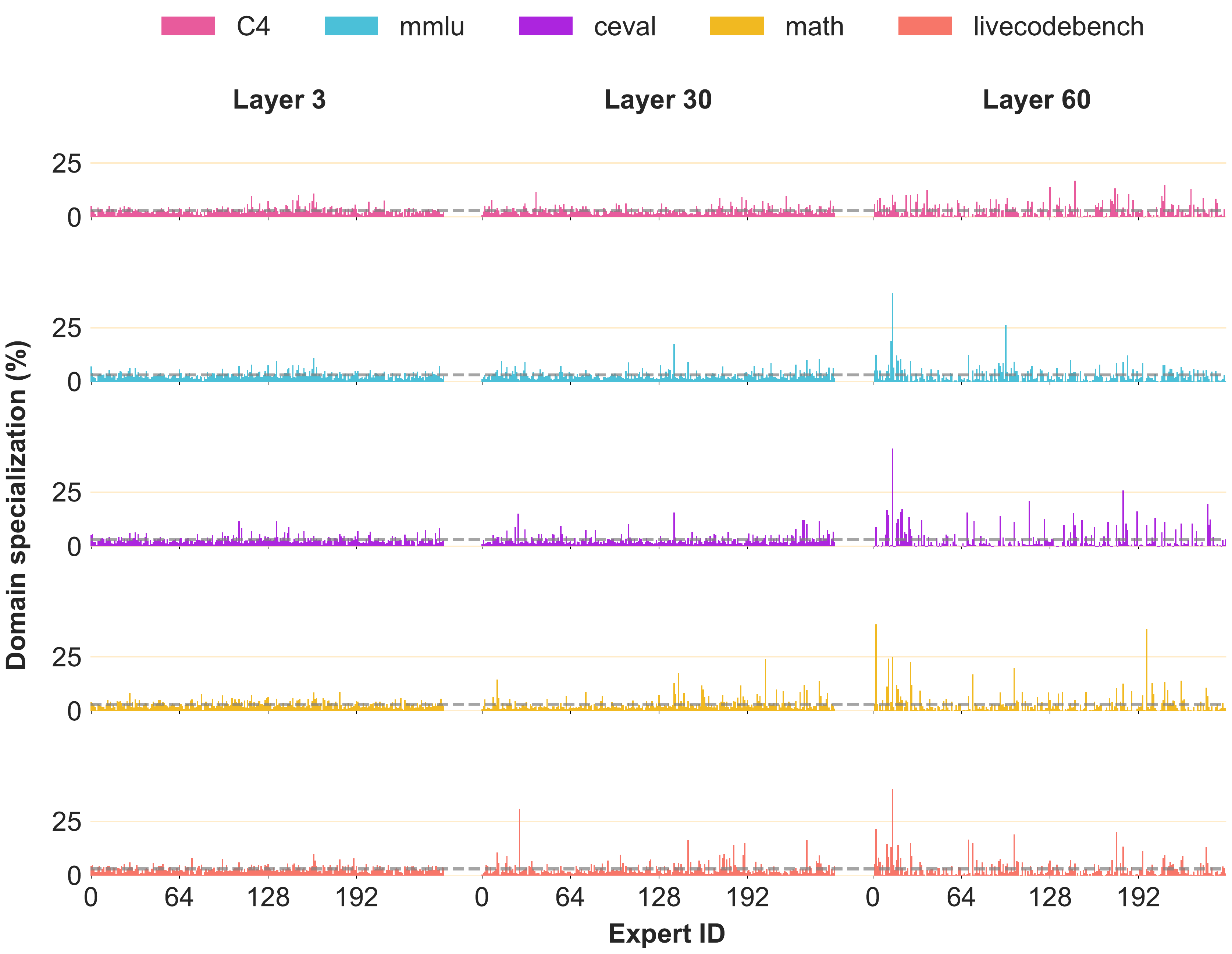}
    \caption{Expert specialization pattern of \modelname. Each subgraph represents the token distribution in a layer for a given task. Each bar within the subgraph corresponds to the percentage of tokens assigned to a specific expert relative to the total number of tokens. \modelname~contains 256 experts per layer, activating 8 experts for each token. Therefore, the average token percentage per expert is 8/256 = 3.13\%, denoted by the gray dotted line. The token distributions across different experts vary significantly, indicating that \modelname~exhibits substantial expert specialization, which contributes to improved training and performance.}
    \label{fig:routing_Pangu Ultra MOE}
\end{figure*}

\minisection{Domain Specialization} The pattern of expert specialization serves as a critical indicator of an MoE layer, as it provides insight into the extent to which the MoE experts have effectively acquired knowledge from the data. In this section, we investigate the phenomenon of expert specialization across various tasks to understand how this pattern is distributed among them. The analysis is conducted using data from five distinct datasets: C4, MMLU, C-Eval, Math, and LiveCodeBench.

As illustrated in Figure~\ref{fig:routing_Pangu Ultra MOE}, we examine the tokens assigned to experts in the shallow, middle, and deep layers (\ie, Layer 3, 30, and 60). In the context of different tasks, tokens at the same layer are preferentially routed to different experts, resulting in significant variability in expert specialization across tasks. Notably, experts in Layer 60 exhibit a higher degree of specialization compared to those in Layer 30, which, in turn, show greater specialization than those in Layer 3. This observation suggests that expert specialization intensifies as the layer depth increases. Furthermore, in Layer 60, the front experts are predominantly favored in most tasks, with the exception of C4, where expert specialization is notably lower compared to the other datasets.

Through the analysis of expert specialization, we demonstrate that \modelname~has developed significant variation in expert specialization, which enhances the model's expressive capability and contributes to its overall performance.

\begin{figure*}[t]
    \centering
    \includegraphics[width=0.8\textwidth]{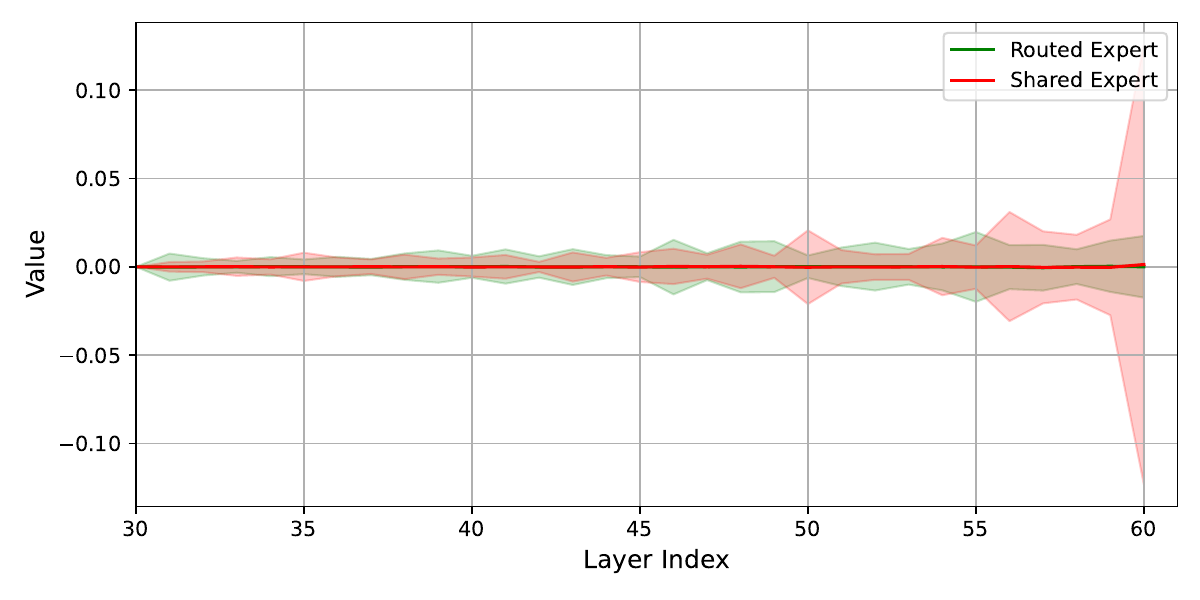}
    \caption{The outputs of routed and shared experts. We display the outputs after the 30th layer for better clarity, as the earlier outputs are small. The mean values of the outputs from both the routed and shared experts approach zero. The standard deviations across the layers are similar, with the exception of the final layer. Overall, the contributions of the routed experts are comparable to those of the shared experts across the layers.}
    \label{fig:output_expert}
\end{figure*}

\minisection{Router Scale}
The output of the MoE layer is the sum of the outputs from both shared and routed experts. Therefore, it is crucial to maintain the balance between these two components, especially the routed experts. Since routed experts constitute a significant portion of the entire model, their failure to contribute meaningfully can severely degrade the performance of the MoE layer, reducing it to the level of a dense layer. To address this, we analyze the outputs of both components to assess the contribution of the routed experts.

As shown in Figure~\ref{fig:output_expert}, the mean values of the outputs from the routed and shared experts are nearly identical across all layers. The standard deviations are comparable in the early layers, with the shared expert's standard deviation becoming larger in the final layer. Overall, the routed experts contribute similarly to the shared experts across the layers, thereby enhancing the representation power of the entire model.

\begin{figure*}[h]
    \centering
    \includegraphics[width=0.7\textwidth]{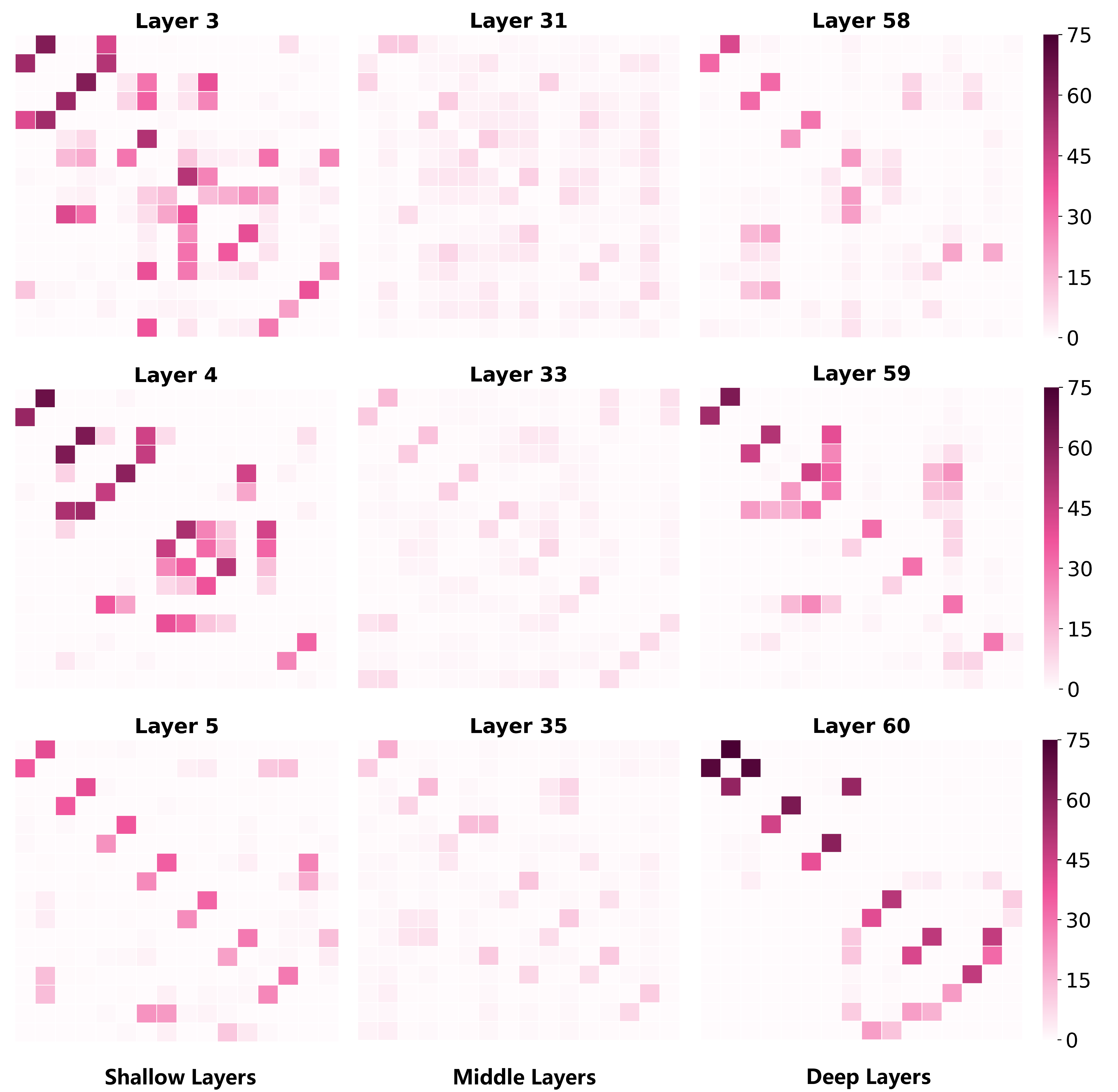}
    \caption{Co-activation among experts in nine layers on a random 0.5\% of the C4 validation data. We display 16 experts with the highest maximum co-activation score via their expert IDs on the x- and y-axis.}
    \label{fig:co_act}
\end{figure*}

\minisection{Expert Co-Activation} We also visualize the expert co-activation situation based on the co-activation matrix. The co-activation score of the matrix represents the probability of one expert being activated and another expert being activated as well. The higher the co-activation score, the greater the correlation between the two experts. 

In order to comprehensively reflect the situation, we select six different layers to represent the front, middle, and back parts of the model. Figure~\ref{fig:co_act} shows there is no strong co-activation among experts in three layers, except for a few exceptions. This may reflect the low redundancy of our experts. What is more, compared to the front and back layers, the co-activation scores of the middle layers are much lower. This may reflect the process of knowledge flow diverging and then converging in the whole model.

\section{Conclusion}
We present a systematic training recipe to efficiently train large-scale sparse MoE models on Ascend NPUs. To address system challenges posed by trillion-parameter models, we propose a simulation-driven strategy for optimizing model hyperparameters, reducing the need for costly hardware experiments. Our system optimizations focus on Expert Parallelism and memory management, significantly lowering communication and activation overhead across 6K NPUs. These innovations enable a 30.0\% MFU, demonstrating Ascend NPUs' capability to support full-scale training of large-scale sparse LLMs, e.g., ~\modelname, with comparable performance as DeepSeek R1. Experiments validate that our method achieves hardware-aligned performance gains while maintaining training stability. Furthermore, our analysis of model behaviors provides insights for future research on balancing computational efficiency and model capacity. This work establishes a practical foundation for deploying massive MoE models in Ascend NPUs.

\bibliographystyle{plain}
\bibliography{ref}  

\newpage
\appendix

\section{Contributions and Acknowledgments}

\noindent
\textbf{Core Contributors} \\
Yehui Tang, Yichun Yin, Yaoyuan Wang, Hang Zhou, Yu Pan, Wei Guo, Ziyang Zhang, Miao Rang, Fangcheng Liu, Naifu Zhang, Binghan Li, Yonghan Dong, Xiaojun Meng, Yasheng Wang, Dong Li, Yin Li, Dandan Tu,  Can Chen, Youliang Yan, Fisher Yu, Ruiming Tang, Yunhe Wang

\textbf{Contributors} \\
Botian Huang, Bo Wang, Boxiao Liu, Changzheng Zhang, Da Kuang, Fei Liu, Gang Huang, Jiansheng Wei, Jiarui Qin, Jie Ran, Jinpeng Li, Jun Zhao, Liang Dai, Lin Li, Liqun Deng, Peifeng Qin, Pengyuan Zeng, Qiang Gu, Shaohua Tang, Shengjun Cheng, Tao Gao, Tao Yu, Tianshu Li, Tianyu Bi, Wei He, Weikai Mao, Wenyong Huang, Wulong Liu, Xiabing Li, Xianzhi Yu, Xueyu Wu, Xu He, Yangkai Du, Yan Xu, Ye Tian, Yimeng Wu, Yongbing Huang, Yong Tian, Yong Zhu, Yue Li, Yufei Wang, Yuhang Gai, Yujun Li, Yu Luo, Yunsheng Ni, Yusen Sun, Zelin Chen, Zhe Liu, Zhicheng Liu, Zhipeng Tu, Zilin Ding, Zongyuan Zhan

\newpage

\section{Case Analysis}
In this section, we conduct a case study, covering finance, medicine and general domains,  to further demonstrate the advantages of \modelname{}. 

Table~\ref{tab:medical-case-1} and~\ref{tab:medical-case-2} show examples from medicine exam questions. \modelname{} only not equips with fluent and decent reasoning capability with medical knowledge, but also leverages internal knowledge from  medical textbook (\eg, \begin{CJK*}{UTF8}{gkai}内科学\end{CJK*}) and medical practice guidelines (\eg, KDIGO 2021), showing the superior medical capability of \modelname{}.

Table~\ref{tab:financial-case-1} and~\ref{tab:financial-case-2} show examples of financial exam questions. \modelname{} can also cite and use knowledge from financial industry documents (\ie, \begin{CJK*}{UTF8}{gkai}《商业银行授信工作尽职引导》\end{CJK*}) in the reasoning process. In addition, \modelname{} can still properly conduct mathematical calculations under complex financial conditions.

Table~\ref{tab:aigc-case-1} and~\ref{tab:aigc-case-2} show examples from general domains. Compared to Deepseek R1, \modelname{} provides more details in the generated plan and document. When answering general knowledge question, \modelname{} produces richer documents with textual tables and clarification questions to ensure the diversity of answer.

\begin{CJK*}{UTF8}{gkai}
\begin{table*}[ht]
  \centering
  \caption{This is a medical case.}
  \resizebox{0.96\textwidth}{!}{
    \begin{tabular}{p{26.5em}l}
    \toprule
    \multicolumn{2}{l}{治疗泄泻寒邪内盛证，首选的方剂是\hspace{0.1cm}A、藿香正气散\hspace{0.1cm}B、理中丸\hspace{0.1cm}C、参苓白术散\hspace{0.1cm}D、四神丸\hspace{0.1cm}E、痛泻要方} \\
    \midrule
    \multicolumn{2}{p{55em}}{\textbf{DeepSeek R1}:
    
    好吧，我现在得仔细看看这个问题。题目是治疗泄泻寒邪内盛证，首选的方剂是哪一个选项。选项有A到E五个，我需要一个一个分析。

首先，泄泻也就是腹泻，寒邪内盛证，说明是寒邪比较严重，导致腹泻。这时候应该用温中散寒的方法来治疗。那各个选项对应的方剂是什么呢？

A选项是藿香正气散，这个我记得是用于外感风寒、内伤湿滞的情况，比如有呕吐、腹泻，同时可能有发热恶寒的表证。藿香正气散主要解表化湿，理气和中，所以如果是外感引起的肠胃不适可能用这个。但题目里提到的是寒邪内盛，可能没有表证，所以这个是不是不太对？

B选项是理中丸，理中丸是温中祛寒，补气健脾的。主治脾胃虚寒证，比如腹痛泄泻，呕吐，手脚不温，舌淡苔白等。寒邪内盛可能属于脾胃虚寒，所以理中丸可能适用。不过需要确认是虚寒还是实寒。如果是寒邪侵袭，属于实寒的话，可能需要其他方剂？

C选项是参苓白术散，这方子是益气健脾，渗湿止泻，用于脾虚湿盛证，比如乏力、食少、便溏，可能偏于脾虚有湿，但这里寒邪内盛，可能不是这个。

D选项是四神丸，主治肾阳虚的五更泄泻，温肾暖脾，涩肠止泻。比如黎明前腹泻，腹痛肠鸣，这种情况属于肾阳虚，所以四神丸可能不是针对寒邪内盛，而是肾阳不足导致的。

E选项是痛泻要方，主要治疗肝气乘脾的痛泻，特点是肠鸣腹痛，泻后痛减，属于肝脾不和，所以这里应该不是。

那回到问题，寒邪内盛导致的泄泻，是实寒还是虚寒？如果是寒邪直中脾胃，导致腹泻，这时候可能需要温中散寒。理中丸针对的是脾胃虚寒，但如果寒邪盛，可能属于实证。不过有时候虚寒和实寒在治疗上可能都用温中的方法。比如理中丸也可以用于中焦寒盛的情况。

或者，如果寒邪内盛，伴有表证，比如恶寒发热，可能用藿香正气散。但题目中没有提到表证，所以可能不是。因此，正确选项应该是B理中丸。
    }\\
    \midrule
    \multicolumn{2}{p{55em}}{\textbf{\modelname{}}:

首先，回忆一下各个方剂的主要功效和主治证型。藿香正气散我记得是用于外感风寒、内伤湿滞的情况，比如感冒伴有呕吐、腹泻，舌苔白腻等。这可能和寒邪有关，但需要确认是否是寒邪内盛，还是外寒内湿。

理中丸，这个方剂是温中散寒的，主要治疗脾胃虚寒证，比如腹痛、呕吐、腹泻，四肢不温，舌淡苔白，脉沉细等。这里的关键是脾胃虚寒，可能属于寒邪内盛的一种情况，但需要看是否是实证还是虚证。因为寒邪内盛可能偏实，而理中丸更多用于虚寒。

参苓白术散是健脾益气的，用于脾虚湿盛导致的泄泻，比如腹泻伴有乏力、食欲不振，舌苔白腻等。这可能偏向于脾虚夹湿，而不是寒邪内盛，所以可能不是正确答案。

四神丸用于肾阳不足导致的五更泄泻，即黎明时分的腹泻，伴有腰膝酸冷，属于命门火衰，不能温煦脾土。这属于脾肾阳虚的泄泻，和寒邪内盛可能有关，但更偏向于阳虚而非外感寒邪。

痛泻要方是治疗肝郁脾虚的痛泻，特点是腹痛即泻，泻后痛减，情绪因素诱发，所以和寒邪内盛关系不大。

现在回到问题，泄泻的寒邪内盛证，应该是寒邪直接侵袭肠胃，导致运化失常，出现泄泻清稀，甚至如水样，腹痛肠鸣，可能伴有恶寒发热，但以寒邪为主。这种情况下，藿香正气散是否适用呢？因为藿香正气散用于外感风寒，内伤湿滞，表里同病的情况，如果有外感症状可能更合适。而如果单纯是寒邪内盛，没有表证，可能需要温中散寒的方剂，比如理中丸。

不过，这里可能需要注意证型的区分。寒邪内盛可能属于实证，而理中丸更多用于虚证。或者是否有其他方剂更合适？

再仔细想想，如果是寒邪内盛，可能用附子理中丸更合适，但选项里没有。而理中丸本身是温中健脾，对于中焦虚寒有效。如果是寒邪直中，可能属于实证，可能需要用更偏于散寒的方剂。或者，藿香正气散是否有温中散寒的作用？

藿香正气散的组成包括藿香、紫苏、白芷等解表散寒的药物，还有厚朴、陈皮、大腹皮等理气化湿的药物，茯苓、白术健脾利湿，半夏和胃。所以它主要针对的是外感风寒，内有湿滞的情况，导致泄泻、呕吐等。如果患者有外感症状，比如恶寒发热，头痛，舌苔白腻，那么用藿香正气散是对的。

\textcolor{red}{而如果患者是纯内寒，没有表证，可能更适合理中丸。所以这里需要明确寒邪内盛证的具体表现。根据中医教材，寒邪内盛证的泄泻，属于实证，治疗当散寒化湿，可能用藿香正气散加减。或者如果是中焦寒盛，可能用理中丸？}

\textcolor{red}{这时候可能需要查阅一下具体的证型和治疗方剂。比如《中医内科学》中，泄泻的寒湿内盛证，治疗首选藿香正气散。如果是脾胃阳虚的泄泻，则用理中丸。所以问题中的证型是寒邪内盛，应该属于寒湿内盛，实证，所以选藿香正气散。而理中丸用于脾胃虚寒证，属于虚证。因此，答案应该是A选项藿香正气散。}

不过我之前可能混淆了虚实的问题，这里需要确认。比如，寒邪内盛可能为实证，而藿香正气散用于外感寒邪，内伤湿滞，属于实证。而理中丸用于脾胃虚寒的虚证。所以正确答案应该是A选项藿香正气散。
    }\\
    \bottomrule
    \end{tabular}%
    }
  \label{tab:medical-case-1}%
\end{table*}%
\end{CJK*}
\begin{CJK*}{UTF8}{gkai}
\begin{table*}[ht]
  \centering
  \scriptsize
  \caption{This is a medical case.}
    \begin{tabular}{p{26.5em}l}
    \toprule
    \multicolumn{2}{l}{慢性肾炎患者尿蛋白 ≥1g/d 时控制高血压的目标值是\hspace{0.1cm} A $\leq$ 150/95\hspace{0.1cm}B $\leq$ 140/90\hspace{0.1cm}C$\leq$ 135/85\hspace{0.1cm}D$\leq$ 130/80\hspace{0.1cm}E$\leq$ 125/75} \\
    \midrule
    \multicolumn{2}{p{60em}}{\textbf{DeepSeek R1}:
    
嗯，我现在得仔细想想这个问题。题目是慢性肾炎患者尿蛋白≥1g/d时，控制高血压的目标值是多少。选项有A到E五个。首先，我需要回忆一下相关的医学知识。

慢性肾炎患者如果有较高的尿蛋白，通常意味着肾脏损害比较严重，所以控制血压的目标可能更严格，以减缓肾脏病的进展。一般来说，普通高血压患者的目标可能是140/90 mmHg以下，但对于有肾脏疾病，尤其是蛋白尿的患者，可能需要更低的血压目标。

我记得在指南里，比如KDIGO（国际肾脏病组织）的指南，推荐对于有蛋白尿的慢性肾病患者，血压应控制在130/80 mmHg以下。特别是当尿蛋白≥1g/d时，可能需要更严格的控制。所以可能选项D是正确答案，即<130/80mmHg。

不过，有时候可能会有不同的指南推荐。比如以前的JNC8可能有些不同，但近年来更倾向于更严格的控制，尤其是有并发症的情况。比如糖尿病患者或者有蛋白尿的患者，血压控制更严格。所以应该确认一下最新的指南。

另外，选项E是<125/75mmHg，这可能过于严格，除非有特别的情况，但通常这个数值可能太低，容易导致低血压的风险，尤其是老年患者。而选项D是130/80，比较常见于推荐中。所以综合考虑，正确答案应该是D选项。
    }\\
    \midrule
    \multicolumn{2}{p{60em}}{\textbf{\modelname{}}:

首先，我记得慢性肾炎患者通常需要控制血压，以减缓肾功能的恶化。而尿蛋白的量是一个重要的指标，可能影响血压控制的目标值。一般来说，对于有蛋白尿的患者，尤其是尿蛋白较多的情况下，血压控制可能更为严格，以减少肾脏的负担。

接下来，我需要回顾一下高血压管理的指南，特别是针对慢性肾脏病（CKD）患者的推荐。根据KDIGO（改善全球肾脏病预后组织）指南，对于CKD患者，尤其是存在蛋白尿的情况，推荐将血压控制在较低的水平。比如，当尿蛋白排泄量较高（比如≥1g/d）时，目标血压可能更低，通常建议收缩压（SBP）<130 mmHg，舒张压（DBP）<80 mmHg。这可能对应选项D，即<130/80 mmHg。

不过，我还需要确认其他可能的来源是否有不同的建议。比如，美国心脏协会（AHA）或欧洲心脏病学会（ESC）的指南是否有不同的标准。过去的一些指南可能推荐更宽松的目标，比如<140/90 mmHg，但对于有肾脏疾病的患者，尤其是伴有蛋白尿时，会更严格。比如，有些指南可能建议在尿蛋白≥1g/d的情况下，目标血压是<125/75 mmHg，这对应选项E。但这是否正确呢？

这里可能存在一些变化，因为不同的指南可能有不同的推荐，或者随着时间的推移，推荐的目标值有所调整。比如，较早的指南可能更倾向于严格控制，但后来的研究可能显示过于严格的控制可能没有额外的好处，甚至可能有风险。比如，SPRINT试验显示，对于高危患者，强化血压控制（SBP<120 mmHg）可能有益，但该试验排除了CKD患者，所以可能不适用这里的情况。

再查一下最新的KDIGO指南，2021年的更新建议对于CKD患者，无论是否有蛋白尿，目标SBP应<120 mmHg，但这是基于标准化诊室血压测量。不过这可能适用于使用标准化测量方法的情况下。而如果尿蛋白较高（如≥1g/d），可能需要更严格的控制，但可能不同的机构有不同的建议。

不过，这个问题可能基于较旧的标准。例如，之前的指南可能推荐对于有蛋白尿的CKD患者，目标血压<130/80 mmHg，而如果尿蛋白≥1g/d，则可能更低，比如<125/75 mmHg。但需要确认这一点是否正确。

另一个角度是，尿蛋白≥1g/d通常意味着更严重的肾脏损害，因此需要更严格的血压控制以减少蛋白尿和保护肾功能。一些研究显示，较低的血压目标（如<130/80 mmHg）可能有助于减少蛋白尿和延缓肾病进展。然而，是否有足够的证据支持更低的目标，比如E选项的<125/75 mmHg？

比如，ACCORD试验研究了糖尿病患者的血压控制，发现强化控制（SBP<120 mmHg）并未显著改善主要心血管结局，但减少了某些次要结局。不过对于肾脏疾病患者，特别是伴有大量蛋白尿的情况，可能有所不同。

再回到问题本身，选项中E是<125/75 mmHg，而D是<130/80 mmHg。根据一些教材或考试的标准答案，可能倾向于E选项，尤其是当尿蛋白超过1g/d时。例如，传统的教学可能指出，对于尿蛋白>1g/d的慢性肾病患者，血压目标应更严格，如<125/75 mmHg，而尿蛋白较低的患者可能目标为<130/80 mmHg。但需要确认这一点是否正确。

\textcolor{red}{不过，现在可能存在不同的意见。例如，KDIGO 2012指南建议，对于CKD患者伴尿蛋白≥1g/d，目标血压应≤130/80 mmHg。而更新的指南可能有所调整。但如果是考试题目，可能仍然以旧的标准答案为参考。}

或者，可能根据不同的尿蛋白水平设定不同的目标。例如，当尿蛋白≥1g/d时，更严格的目标，如<125/75 mmHg，而如果尿蛋白较低，则<130/80 mmHg。这种情况下，正确答案可能是E选项。

\textcolor{red}{不过，我需要再次核实。比如，查阅UpToDate或其他可靠资源。根据UpToDate关于慢性肾炎的治疗，对于高血压的管理，当尿蛋白排泄量≥1g/d时，推荐将血压控制在≤130/80 mmHg，如果能够耐受，甚至更低。但可能不会明确推荐到125/75 mmHg。这可能意味着选项D是正确的。}

但有些资料可能指出，对于大量蛋白尿患者（如>3.5g/d，即肾病范围蛋白尿），可能需要更严格的控制，但题目中的情况是≥1g/d，可能属于非肾病性蛋白尿，但仍然较高。

\textcolor{red}{或者，可能根据不同的指南版本。例如，美国糖尿病协会（ADA）和KDIGO的联合报告可能建议对于糖尿病肾病伴尿蛋白排泄增加的患者，目标血压<130/80 mmHg。而对于非糖尿病肾病，同样适用类似的推荐。}

但问题中的患者是慢性肾炎，可能属于非糖尿病肾病。因此，是否应该选择D选项？

或者，是否存在某些情况下，当尿蛋白≥1g/d时，目标更严格？

\textcolor{red}{比如，根据JNC 8指南，对于CKD患者，无论蛋白尿情况，目标血压<140/90 mmHg。这可能对应选项B。但JNC 8的推荐较为宽松，可能与其他指南不同。但JNC 8的发布时间是2014年，而KDIGO的更新可能更严格。}

因此，这题可能存在不同的答案，取决于所参考的指南。但在中国，可能采用的是更严格的标准。例如，中国指南可能推荐对于慢性肾炎伴蛋白尿≥1g/d的患者，血压控制目标为<130/80 mmHg，或者更严格。

不过，根据中国慢性肾脏病高血压管理指南，可能推荐对于伴有蛋白尿的患者，目标血压<130/80 mmHg，若能耐受可更低。因此，选项D是正确答案。

\textcolor{red}{但有些教材可能指出，当尿蛋白≥1g/d时，血压应控制在<125/75 mmHg。例如，在之前的医学教材中，如《内科学》教材，可能提到对于大量蛋白尿患者，更严格的血压控制。因此，可能选项E是正确的。}

这让我有些困惑，因为不同的来源可能有不同的推荐。我需要再仔细回忆或查找相关资料。

\textcolor{red}{例如，人民卫生出版社的第9版《内科学》教材中，关于慢性肾小球肾炎的治疗，提到高血压的治疗目标：尿蛋白≥1g/d时，血压应控制在<125/75 mmHg；尿蛋白<1g/d时，血压控制可放宽到<130/80 mmHg。因此，根据这个教材，当尿蛋白≥1g/d时，正确答案是E选项，即<125/75 mmHg。}

所以，如果这个问题是基于教材内容，那么正确答案应该是E选项。但需要注意到，不同的教材版本可能有不同的推荐，或者在最新的指南中可能已经更新。

\textcolor{red}{例如，KDIGO 2021指南建议，对于CKD患者，无论是否有蛋白尿，收缩压目标为<120 mmHg，但这是基于标准化诊室血压测量。然而，在临床实践中，可能仍存在不同的应用，尤其是在考试中可能仍沿用旧的标准。}

因此，结合中国教材的内容，正确答案应为E选项，即<125/75 mmHg。
    }\\
    \bottomrule
    \end{tabular}%
  \label{tab:medical-case-2}%
\end{table*}%
\end{CJK*}
\begin{CJK*}{UTF8}{gkai}
\begin{table*}[ht]
  \centering
  \caption{This is a financial case.}
  \scriptsize
  \resizebox{0.96\textwidth}{!}{
    \begin{tabular}{p{26.5em}l}
    
    \toprule
    \multicolumn{2}{p{100em}}{\fontsize{12pt}{14pt}\selectfont在授信额度批准后，商业银行在具体操作中，还应针对不同授信品种的性质、特点，对客户信用风险以外的风险进行严格审查，统一审核发放。（）\hspace{0.1cm}A. 对\hspace{0.1cm}B. 错} \\
    \midrule
    \multicolumn{2}{p{100em}}{\fontsize{12pt}{14pt}\selectfont\textbf{DeepSeek R1}:
    
嗯，我现在要解决这个问题，题目是关于商业银行在授信额度批准后的操作是否需要针对不同授信品种的风险进行严格审查，并统一审核发放。选项是A对，B错。首先，我得回忆一下相关的银行风险管理知识。

首先，授信额度批准后，银行是不是还需要进一步审查其他风险呢？题目中提到的是“客户信用风险以外的风险”，比如可能有市场风险、操作风险、法律风险等等。根据巴塞尔协议或者商业银行风险管理的一般流程，银行在授信之后，实际发放贷款或提供信用时，应该考虑不同授信产品的特性，因为这些产品可能有不同的风险因素。例如，短期贷款和长期贷款的风险不同，信用证和保函也有不同的风险点。

题目中说“还应针对不同授信品种的性质、特点，对客户信用风险以外的风险进行严格审查，统一审核发放。”这里的“统一审核发放”是什么意思呢？统一审核是否指所有授信品种都要经过同样的审核流程，还是说根据不同的品种有不同的审核步骤？或者可能是指审核的标准要统一，但具体到不同产品时，侧重点不同？

我记得商业银行在发放贷款或其他授信时，虽然授信额度已经批准，但具体到每一笔业务，还需要进行具体的审查，包括用途、合规性、其他风险因素。比如，即使客户有一个总的授信额度，每次提款时银行可能还要检查当时的市场状况、客户的财务状况是否有变化，以及该笔具体交易的风险。

另外，题目提到的是“客户信用风险以外的风险”，也就是除了客户的还款能力和意愿之外的其他风险，比如交易结构风险、抵押品风险、国家风险、法律风险等。这些风险可能因不同的授信产品而不同。例如，国际贸易融资涉及汇率风险和国别风险，而房地产贷款则涉及抵押品价值和市场风险。

所以，银行在具体操作中确实需要针对不同授信品种的特点，审查其他相关风险，并且可能需要统一审核流程，确保所有风险都被适当评估后再发放。因此，题目中的说法应该是对的，即选项A正确。

不过，也有可能题目中的“统一审核发放”表述有问题。是否所有授信品种的审核都应该统一，还是根据不同品种有不同的审核程序？如果“统一”指的是统一的标准和流程，但具体执行时考虑不同产品的特性，那么是对的。如果“统一”意味着不考虑差异，统一处理，那可能是不对的。但题目中的后半句是“针对不同授信品种的性质、特点”，所以可能是在统一的标准下，根据不同产品进行调整，因此审查是严格且统一的流程，但内容上有所侧重。

综合来看，正确的答案应该是A，对。因为即使在授信额度批准后，实际发放时仍需审查其他风险，并根据产品特性进行调整，同时保持审核流程的统一性。
    }\\
    \midrule
    \multicolumn{2}{p{100em}}{\fontsize{12pt}{14pt}\selectfont\textbf{\modelname{}}:

嗯，我现在要解决这个题目，关于商业银行在授信额度批准后的操作流程。题目是说，商业银行在具体操作中，应该针对不同授信品种的性质和特点，对客户信用风险以外的风险进行严格审查，统一审核发放。然后问这个说法是对还是错，选项是A对，B错。

首先，我需要回忆一下商业银行授信流程的相关知识。授信额度批准后，银行确实需要进行一系列的风险审查，但这里题目提到的是“信用风险以外的风险”，比如市场风险、操作风险、法律风险等。\textcolor{red}{根据巴塞尔协议，银行需要全面管理各种风险，不仅仅是信用风险。}

然后，题目中提到“统一审核发放”，这可能有问题。因为在实际操作中，不同的授信品种可能有不同的审核流程和标准。例如，贷款和信用证虽然都属于授信，但它们的风险点不同，处理流程也可能不同。统一审核可能不太准确，应该是个别审核，根据具体业务类型来分别处理。

不过，也有可能题目中的“统一审核”指的是在某个统一的框架或标准下进行审核，而不是说所有品种都完全一样的流程。\textcolor{red}{这时候需要判断这里的“统一”是否正确。根据中国银监会的《商业银行授信工作尽职指引》，银行应对不同授信业务进行分门别类的管理，所以可能“统一审核”不正确，应该是分别审核。}

所以，题目中的说法是否正确呢？如果严格审查信用风险以外的其他风险，并且统一审核发放，是否正确？可能错误，因为审核应该根据不同的授信品种分别进行，而不是统一审核。因此，正确选项应该是B错。
    }\\
    \bottomrule
    \end{tabular}%
    }
  \label{tab:financial-case-1}%
\end{table*}%
\end{CJK*}
\begin{CJK*}{UTF8}{gkai}
\begin{table*}[ht]
  \centering
  \caption{This is a financial case.}
  \resizebox{0.96\textwidth}{!}{
    \begin{tabular}{p{26.5em}l}
    \toprule
    \multicolumn{2}{p{120em}}{\fontsize{20pt}{21pt}\selectfont甲公司2016年初流通在外普通股8000万股，优先股500万股；2016年6月30日增发普通股4000万股。2016年末股东权益合计35000万元，优先股每股清算价值10元，无拖欠的累积优先股股息。2016年末甲公司普通股每股市价12元，市净率是（）\hspace{0.1cm}A. 2.8\hspace{0.1cm}B. 4.8\hspace{0.1cm}C. 4\hspace{0.1cm}D. 5} \\
    \midrule
    \multicolumn{2}{p{120em}}{\fontsize{20pt}{21pt}\selectfont\textbf{DeepSeek R1}:
    
好，我现在要解这道题，题目是关于市净率的计算。首先我得回忆一下市净率的定义，应该是每股市价除以每股净资产对吧？嗯，对的，市净率=每股市价/每股净资产。题目里已经给出每股市价是12元，所以关键是要找到每股净资产是多少，然后用12除以这个数就能得到市净率了。

接下来我得仔细看一下题目中的数据。题目里说甲公司2016年初流通在外的普通股是8000万股，优先股500万股；然后在2016年6月30日增发了普通股4000万股。2016年末的股东权益合计是35000万元，优先股每股清算价值10元，而且没有拖欠的累积优先股股息。最后问市净率是多少，选项是A到D四个。

首先，我需要计算普通股的每股净资产。这里可能需要注意普通股和优先股的权益分配。首先，股东权益合计是35000万元，其中优先股的权益应该先扣除，剩下的才是普通股的权益。

优先股的清算价值是每股10元，有500万股优先股，所以优先股的清算价值总额就是500万×10元=5000万元。题目里还提到没有拖欠的累积优先股股息，所以不需要考虑应付股息的问题，只需要清算价值的话，那普通股的权益就是股东权益合计减去优先股的清算价值，也就是35000万 - 5000万=30000万元。

接下来计算普通股的股数。这里要注意时间权数，因为年中增发了新股。2016年初有8000万股普通股，6月30日增发4000万股，也就是下半年才有的。所以普通股的加权平均股数应该是：

从1月1日到6月30日，共6个月，有8000万股；从7月1日到12月31日，也是6个月，有8000万+4000万=12000万股。所以加权平均股数=8000万×(6/12) + 12000万×(6/12)= 4000万+6000万=10000万股。或者也可以直接用时间加权计算，比如增发是在年中，所以相当于全年有8000万加上4000万的半年的部分，即4000万×0.5=2000万，所以总股数就是8000+2000=10000万股。两种方法应该都对，结果都是10000万股。

那么普通股的每股净资产就是普通股权益除以普通股加权平均股数，也就是30000万/10000万=3元。然后市净率就是每股市价12元除以每股净资产3元，得到12/3=4，所以答案应该是C选项4。

不过，我得再检查一下步骤有没有哪里出错了。首先，股东权益合计是35000万，扣除优先股清算价值5000万，剩下的30000万属于普通股。对吧？是的，优先股的清算价值先于普通股分配，所以这部分要先减掉。然后计算普通股的加权平均股数，确实要考虑增发的时间，年中增发的话，相当于后半年的股数增加，所以加权平均是10000万股。然后30000万除以10000万就是3元的每股净资产，市净率12/3=4，选项C正确。

不过，有没有可能哪里考虑得不对？比如，优先股清算价值和账面价值的问题，或者普通股权益是否需要考虑其他项目？题目里没有提到优先股的账面价值，而是直接给了清算价值，所以应该用清算价值来计算，而不是账面价值，对吗？

是的，题目里说的清算价值是每股10元，而清算的时候优先股的权益就是按照清算价值计算的，所以这里应该扣除的是清算价值，也就是500万×10=5000万，这样普通股的权益就是35000万-5000万=30000万。这部分是对的。

另外，普通股的加权平均股数计算，题目里是2016年6月30日增发4000万股，所以从7月1日开始有这4000万股，时间权重是半年，所以应该是4000万×6/12=2000万股，加上原来的8000万全年都是吗？不，原来的8000万在增发前半年是8000万，后半年也是8000万加上4000万，所以正确的计算应该是：

前半年（6个月）的股数：8000万，时间权重是6/12，所以贡献的加权股数是8000×0.5=4000万。

后半年（6个月）的股数是8000+4000=12000万，时间权重也是6/12，所以贡献的是12000×0.5=6000万。

总加权平均股数=4000+6000=10000万，没错。

所以每股净资产3元，市净率12/3=4，选项C。看起来是对的，不过再仔细看看题目里有没有其他条件或者可能忽略的地方。

题目里说“2016年末股东权益合计35000万元”，所以这里是不是已经包含了增发普通股带来的权益增加？比如，增发4000万股，如果按面值或者其他价格发行，是否会影响股东权益？不过题目里没有给出发行价格，只是说增发了普通股，这部分可能已经被包含在年末的股东权益合计35000万里面了，所以在计算普通股权益的时候不需要额外调整，因为增发已经反映在35000万里面了。对吗？

是的，股东权益合计是年末的总额，已经包括了所有普通股和优先股的权益，包括增发的部分。所以不需要另外计算增发带来的权益，只需要从总权益中扣除优先股的部分，剩下的就是普通股权益，对吗？对的，所以这个过程是对的。

所以最终答案应该是选项C，4。不过，题目里选项中有没有这个选项呢？选项C是4，对的。所以应该选C。
    }\\
    \midrule
    \multicolumn{2}{p{120em}}{\fontsize{20pt}{21pt}\selectfont\textbf{\modelname{}}:

首先，市净率的计算公式是每股市价除以每股净资产。我需要先计算每股净资产，然后用市价12元除以它得到市净率。

不过，这里涉及到普通股和优先股的问题，还有股东权益的分配，所以我得仔细处理。

首先，股东权益合计是35000万元，这部分应该包括普通股和优先股的权益。但题目中提到优先股的清算价值是每股10元，而且没有拖欠的累积股息，所以优先股的权益应该是清算价值乘以股数，也就是500万股×10元=5000万元。剩下的股东权益属于普通股，也就是35000万元减去5000万元，等于30000万元。

接下来，我需要计算普通股的流通在外的平均股数还是年末的股数？这里市净率是期末每股市价除以期末每股净资产，所以应该用期末的股数来计算每股净资产。

不过，题目中的普通股变动是在2016年6月30日增发了4000万股，也就是在年中增发。但是计算每股净资产的时候，是否需要考虑时间权重呢？这点需要明确。

通常，每股净资产（Book Value per Share）是按照期末的股数来计算的，因为净资产是期末的时点数，所以股数也应该用期末的股数，而不需要考虑加权平均。比如，期初有8000万股，年中增发4000万，到年末总共有8000+4000=12000万股普通股。因此，计算每股净资产的时候，普通股股数是12000万股。

那普通股的股东权益是35000万减去优先股的5000万，即30000万。所以每股净资产就是30000万除以12000万股，等于2.5元/股。然后市价是12元，市净率就是12 / 2.5 = 4.8，对应选项B？

不过，我可能哪里出错了，因为选项里有选项B是4.8，但让我再仔细检查一遍。

首先，股东权益合计35000万，优先股清算价值是每股10元，共500万股，所以优先股部分总额是500万×10=5000万。没错，剩下的普通股权益是35000万-5000万=30000万。

然后普通股的股数，期初8000万，年中增发4000万，到年末总共有8000+4000=12000万股。没错，所以每股净资产是30000万 / 12000万=2.5元。市价12元，市净率12/2.5=4.8，即选项B。对吗？

不过，可能有人会疑问，计算普通股权益的时候是否需要考虑优先股股息？题目里说无拖欠的累积优先股股息，所以可能不需要扣除股息，因为已经支付了或者没有累积拖欠，所以普通股权益就是股东权益合计减去优先股的清算价值。所以我的计算是对的。

所以答案应该是B选项4.8。不过，我是不是哪里考虑错了？

另一个可能的错误点：计算每股净资产的时候是否需要使用加权平均股数？例如，在计算每股收益的时候，通常用加权平均股数，但市净率是否也如此？

这里需要明确，每股净资产是期末股东权益除以期末股数，而不是加权平均。因为股东权益是时点数，对应的股数也是期末的股数，所以不需要加权平均。比如，如果公司增发股票，那么增发后的股数会增加，同时股东权益也会增加，所以计算每股净资产时，直接使用期末的股数。

例如，假设年初有8000万股，6月30日增发4000万股，那么年末股数是12000万股。这部分股数在计算每股净资产时，直接使用12000万。所以前面的计算是对的，每股净资产2.5元，市净率12/2.5=4.8，选项B。对吗？

不过，有些教材可能会有不同的处理方式，但根据常规做法，应该是对的。所以答案选B，4.8。

不过，再仔细看一下题目，题目中的股东权益合计是35000万元，是否已经扣除了优先股的部分？或者是否应该从股东权益中减去优先股的清算价值？

是的，通常股东权益总额包括普通股和优先股，但计算普通股的每股账面价值时，需要将优先股的部分从股东权益中扣除。这里题目已经给出优先股的清算价值是每股10元，500万股，所以优先股总权益是5000万元，剩下的30000万元属于普通股。这部分处理是对的。

所以，普通股每股净资产=30000万元 / 12000万股=2.5元/股，市价12元，市净率=12/2.5=4.8，选B。没错，应该选B。

不过，我之前好像看到有的题目可能要考虑优先股的股息，但这里题目说“无拖欠的累积优先股股息”，所以不需要考虑应付未付的股息，所以优先股的权益就是清算价值5000万元，对吗？

是的，清算价值是优先股在清算时的价值，如果没有拖欠股息，那么优先股股东权益就是清算价值，所以没错。

总结一下：

1. 计算优先股权益：500万 ×10元=5000万元。

2. 普通股权益=35000万 -5000万=30000万元。

3. 普通股股数：年初8000万，年中增发4000万，年末12000万。

4. 普通股每股净资产=30000万 /12000万=2.5元。

5. 市净率=市价12元 / 每股净资产2.5元=4.8，选项B。

所以答案应该是B选项4.8。
    }\\
    \bottomrule
    \end{tabular}%
    }
  \label{tab:financial-case-2}%
\end{table*}%
\end{CJK*}

\begin{CJK*}{UTF8}{gkai}
\begin{table*}[ht]
  \centering
  \caption{This is an AIGC case.}
  \scriptsize
  \resizebox{0.96\textwidth}{!}{
    \begin{tabular}{p{26.5em}l}
    \toprule
    \multicolumn{2}{l}{去人大招聘实习生，需要注意哪些？} \\
    \midrule
    \multicolumn{2}{p{55em}}{\textbf{DeepSeek R1}:

去中国人民大学（人大）招聘实习生时，需要注意以下几个方面，以确保招聘效果并吸引优质人才：

**一、前期准备**

1. **明确岗位需求**  
   - 针对人大学生的专业优势（如法学、经济学、公共管理、新闻传播等）设计岗位，明确实习内容、技能要求、工作时间和薪资待遇。
   - 突出企业品牌和实习生的成长机会（如培训体系、转正通道）。

2. **联系校方资源**  
   - 提前与人大就业指导中心或相关院系对接，确认校招流程、场地安排和时间节点。
   - 通过校内公众号、院系公告栏、学生社群等渠道推广招聘信息。

3. **准备宣传材料**  
   - 制作简洁清晰的招聘海报、宣传单页、公司介绍PPT等，突出企业核心竞争力和实习生福利。
   - 若有校友资源，可邀请人大毕业的员工作为“企业代言人”参与宣讲。

**二、招聘流程设计**
1. **宣讲会或双选会**  
   - 组织线下/线上宣讲会，介绍企业文化、业务方向、实习生培养计划，并预留互动问答时间。
   - 可设置“快闪面试”环节，现场接收简历并初步筛选。

2. **简历筛选与面试**  
   - 人大学生普遍综合素质较高，需注重简历中的学术背景、实践经历和技能匹配度。
   - 面试时侧重考察逻辑思维、学习能力、抗压能力，避免过于强调经验（实习生更看重潜力）。

3. **灵活考核方式**  
   - 可设置案例分析、小组讨论等环节，观察候选人的团队协作和问题解决能力。
   - 对技术类岗位可增加笔试或实操测试。

**三、吸引人大学子的关键点**

1. **职业发展价值**  
   - 强调实习与未来职业发展的关联性，例如参与核心项目、接触行业资源、获得导师指导等。
   - 提供清晰的晋升路径（如转正机会或推荐信）。

2. **匹配学生特点**  
   - 人大学生普遍关注社会价值，可突出企业社会责任、行业影响力或岗位的公共政策相关性。
   - 灵活的工作时间（如兼顾学业）、交通/餐补等细节需明确说明。

3. **薪资与福利竞争力**  
   - 提供符合市场水平的实习薪资（北京地区需考虑生活成本），或通过非现金福利（如免费培训、活动参与机会）吸引学生。

**四、注意事项**

1. **遵守法律法规**  
   - 签订正规实习协议，明确工作内容、薪资、保密条款等，避免劳动纠纷。
   - 工作时间需符合《劳动法》规定，保护实习生权益。

2. **提升候选人体验**  
   - 及时反馈面试结果，即使拒绝也保持礼貌，维护企业形象。
   - 入职前做好岗前培训，帮助实习生快速融入团队。

3. **长期关系维护**  
   - 建立实习生人才库，对未录用的优秀学生保持联系，未来有岗位时优先推荐。
   - 通过校友网络或企业活动持续扩大在校园的影响力。

**五、案例参考**

- **企业A**：在人大法学院举办“法律实务研讨会”，吸引法学实习生，现场提供模拟法庭案例分析，增强互动。
- **企业B**：联合人大就业中心推出“实习+导师计划”，由高管一对一指导，提升学生参与意愿。

通过以上策略，既能高效匹配优质实习生，也能在人大校园中树立良好的雇主品牌，为未来校招积累资源。
    }\\
    \midrule
    \multicolumn{2}{p{55em}}{\textbf{\modelname{}}:

在中国人民大学（人大）招聘实习生时，为确保招聘效果并吸引优秀学生，需结合学校特点和学生需求，注意以下关键事项：

**一、明确人大优势与学生特点**

1. \textcolor{red}{**学科侧重**：人大以人文社科见长，优势学科包括法学、经济学、管理学、新闻传播、社会学、公共管理等，理工科相对较少。确保岗位与这些领域匹配，或明确交叉学科需求（如数据科学在经济/金融中的应用）。}

2. \textcolor{red}{**学生特质**：人大生源综合素质高，注重理论与实践结合，多数学生对政策研究、智库咨询、金融、法律等领域兴趣浓厚。}

**二、岗位设计与宣传策略**

1. **岗位吸引力**：
   - **内容**：避免“打杂”岗位，突出专业性和成长性（如参与政策分析、课题研究、行业调研等）。
   - **命名**：避免笼统的“实习生”，可细化为“金融分析实习生（宏观经济方向）”“公共政策研究助理”等。
   - **转正机会**：明确是否有留用可能，人大高年级学生（如研二、大三）更关注转正潜力。
   
2. **宣传渠道**：
   - \textcolor{red}{**官方合作**：通过人大就业指导中心官网（http://career.ruc.edu.cn）发布信息，申请线下宣讲会或双选会。}
   - **校内渗透**：联系相关学院（如经济学院、法学院、商学院）的就业辅导员，利用院系微信群、公众号精准推送。
   - **学生社群**：\textcolor{red}{与人大BBS（天地人大）、学生社团（如职业发展协会）合作，或通过校友内推。}
---

**三、招聘流程优化**

1. **时间安排**：
   - \textcolor{red}{**避开考试周**：学期末（6月、12月）学生专注考试，建议在学期中段（4-5月、9-10月）启动招聘。}
   - **反馈效率**：人大生通常投递多家企业，需缩短流程周期（如1周内反馈初筛结果）。

2. **考核方式**：
   - **笔试**：若需考察专业能力，可结合案例分析（如经济类岗位提供政策分析题）。
   - **面试**：建议安排业务部门负责人参与，突出专业深度，避免泛泛而谈。

**四、待遇与法律合规**

1. **实习待遇**：
   - **薪资**：\textcolor{red}{人大位于北京，建议日薪不低于150元（科研助理类岗位可适当调整）。}
   - **福利**：提供餐补、交通补贴或远程办公灵活性，优秀者可承诺推荐信或实习奖励。

2. **协议规范**：
   - 签订《实习协议》，明确工作内容、时长、保密条款及意外保险。
   - 避免要求“强制坐班”，尊重学生课表时间（如每周3天到岗）。

**五、品牌形象与长期合作**

1. **宣讲会设计**：
   - 邀请人大校友分享职业经历，增强学生认同感。
   - 提供行业前沿洞察（如政策解读、金融趋势），吸引学术型学生。

2. **校企合作**：
   - 与学院共建实习基地，赞助学术竞赛或课题研究。
   - 定期开放企业参访，邀请学生参与短期项目（如周末工作坊）。

**六、后续跟进**

- **录用沟通**：明确入职培训计划，发送欢迎邮件并附上团队介绍。

- **反馈收集**：实习结束后向学生和导师征求意见，优化未来招聘策略。

**总结**：\textcolor{red}{人大招聘需注重岗位与学科的匹配度、流程高效性及学生职业发展诉求。通过精准宣传、专业考核和长期合作，可建立优质人才输送渠道。建议提前2-3个月联系校方，预留充足时间协调资源。}
    } \\
    \bottomrule
    \end{tabular}%
    }
  \label{tab:aigc-case-1}%
\end{table*}%
\end{CJK*}

\begin{CJK*}{UTF8}{gkai}
\begin{table*}[ht]
  \centering
  \caption{This is an AIGC case.}
  \scriptsize
  \resizebox{0.96\textwidth}{!}{
    \begin{tabular}{p{26.5em}l}
    \toprule
    \multicolumn{2}{l}{芯片测试领域，SSI是什么维测模式(相较于DFT，design for test）} \\
    \midrule
    \multicolumn{2}{p{55em}}{\textbf{DeepSeek R1}:
    
在芯片测试领域，**SSI**（**System-level Scan Interface**）是一种针对**系统级测试和维护**的维测模式（维护与测试模式），其核心目标是在芯片运行或部署后，通过系统级接口实现功能监测、诊断和调试。相较于传统的**DFT（Design for Test）**技术（如扫描链、BIST等），SSI更侧重于**系统集成后的实时或周期性测试**，而非制造阶段的缺陷检测。

SSI的关键特点与对比DFT：

1. **应用场景不同**：
   - **DFT**：在设计阶段嵌入测试结构（如扫描链、边界扫描JTAG），主要用于芯片制造完成后的**生产测试**（检测制造缺陷）。
   
   - **SSI**：在芯片部署到系统后，通过系统级接口（如高速总线、专用调试接口）进行**运行时的功能验证、性能监控和故障诊断**，属于**运维阶段的测试**。

2. **实现方式差异**：

   - **DFT**：依赖芯片内部预置的测试逻辑（如扫描触发器、BIST引擎），需要占用芯片面积和功耗。
   
   - **SSI**：利用系统已有的通信接口（如PCIe、以太网或专用调试接口），通过协议解析和命令交互实现测试，通常无需额外硬件开销。

3. **测试目标**：

   - **DFT**：主要检测制造相关的物理缺陷（如开路、短路、延迟故障）。
   
   - **SSI**：针对系统级问题，如信号完整性（SI）、电源噪声、软错误（如单粒子翻转）以及软件与硬件的交互故障。

4. **灵活性与动态性**：

   - **DFT**：测试模式固定，需通过ATE（自动测试设备）执行，测试向量在制造阶段固化。
   
   - **SSI**：支持动态配置测试策略，可在系统运行过程中实时调整测试参数，甚至结合软件驱动实现自适应测试。

典型应用场景：

- **复杂SoC的运维监测**：通过SSI实时监控芯片温度、电压、关键信号路径的完整性。

- **车载/航天电子**：在恶劣环境中检测瞬时故障并触发自修复机制。

- **数据中心芯片**：周期性扫描系统状态，预防因老化或软错误导致的宕机。

总结：
SSI是DFT的补充而非替代，两者结合可覆盖芯片全生命周期（从制造到运维）的测试需求。DFT确保芯片出厂质量，而SSI保障其在复杂系统中的长期可靠运行。随着芯片系统复杂度的提升（如AI加速器、自动驾驶SoC），SSI的重要性日益凸显。
    }\\
    \midrule
    \multicolumn{2}{p{55em}}{\textbf{\modelname{}}:

在芯片测试领域，**SSI（System-Level Self-Test Integration）** 是一种**系统级自测试集成**的维测模式，与传统的 **DFT（Design for Test，可测试性设计）** 相比，它更侧重于在系统运行阶段通过软件或固件实现动态测试功能，而非依赖硬件结构的设计优化。以下是两者的对比分析：

**SSI（System-Level Self-Test Integration）**

1. **核心思想**  
   SSI 是在系统级（如芯片集成到终端设备后）通过软件驱动或固件实现的测试模式，利用系统自身资源（如处理器、内存、接口等）执行自测试功能，通常用于运行时故障检测或周期性健康监测。

2. **特点**  
   - **动态测试**：在系统运行过程中实时或按需触发测试，支持在线故障诊断（如老化监测、功能异常检测）。  
   - **软件驱动**：依赖软件算法或固件指令生成测试用例，无需额外硬件电路（或仅需少量硬件配合）。  
   - **系统级覆盖**：可测试芯片与其他系统组件（如传感器、电源管理模块）的交互问题，而不仅是芯片内部结构。  
   - **灵活性高**：测试逻辑可通过软件更新，适配不同应用场景。

3. **典型应用**  
   - 汽车电子中的功能安全（ISO 26262）要求系统实时自检。  
   - 数据中心芯片的运行时健康监测（如温度、电压异常检测）。  
   - 物联网设备的低功耗周期性自测试。

**DFT（Design for Test）**

1. **核心思想**  
   DFT 是在芯片设计阶段插入硬件结构（如扫描链、BIST电路），以提升制造缺陷的检测效率，主要用于芯片量产后的**静态测试**（如ATE测试）。

2. **特点**  
   - **硬件依赖**：需额外设计测试电路（如扫描触发器、MBIST控制器），占用芯片面积和功耗。  
   - **制造阶段测试**：主要用于芯片出厂前的缺陷筛查（如stuck-at故障、延迟故障）。  
   - **结构化测试**：针对预定义的故障模型生成测试向量，覆盖晶体管级或门级缺陷。  
   - **高覆盖率**：通过自动化工具（如ATPG）生成测试模式，确保缺陷检测率。

3. **典型应用**  
   - 芯片量产时的自动化测试（ATE）。  
   - 扫描链测试（Scan Test）、内存BIST（Built-In Self-Test）。  
   - 边界扫描（Boundary Scan）用于PCB互联测试。

**SSI与DFT的关键区别**

\textcolor{red}{| **维度**       | **SSI**                          | **DFT**                          |}

\textcolor{red}{| **测试阶段**   | 系统运行时（动态测试）           | 芯片制造后（静态测试）           |}

\textcolor{red}{| **实现方式**   | 软件/固件驱动，利用系统资源      | 硬件结构设计（扫描链、BIST等）   |}

\textcolor{red}{| **测试目标**   | 功能异常、环境干扰、老化问题     | 制造缺陷（stuck-at、延迟故障等） |}

\textcolor{red}{| **覆盖范围**   | 系统级交互与功能完整性           | 芯片内部结构与制造缺陷           |}

\textcolor{red}{| **灵活性**     | 高（可通过软件更新）             | 低（硬件固化后不可修改）         |}

**互补关系**

- **DFT** 确保芯片出厂时的硬件可靠性，而 **SSI** 保障系统运行时的长期稳定性。 

- 现代芯片设计中，二者常结合使用：DFT用于量产测试，SSI用于现场维护和功能安全（如汽车、航空航天领域）。

\textcolor{red}{若用户提到的SSI是其他缩写（如Scan Segment Insertion），需结合具体上下文进一步解释，但通常SSI在系统级测试语境下指上述模式。}
    } \\
    \bottomrule
    \end{tabular}%
    }
  \label{tab:aigc-case-2}%
\end{table*}%
\end{CJK*}

\end{document}